\documentclass[10pt,twocolumn,letterpaper]{article}

\usepackage{iccv}
\usepackage{times}
\usepackage{epsfig}
\usepackage{graphicx}
\usepackage{amsmath}
\usepackage{amssymb}
\usepackage{multirow}
\usepackage{array}

\usepackage{caption}
\usepackage{bm}
\usepackage[export]{adjustbox} 

\newcommand{\bA}{\mathbf{A}}
\newcommand{\bb}{\mathbf{b}}
\newcommand{\bI}{\mathbf{I}}
\newcommand{\bl}{\mathbf{l}}
\newcommand{\bn}{\mathbf{n}}
\newcommand{\bv}{\mathbf{v}}
\newcommand{\bp}{\mathbf{p}}

\newcommand{\brho}{ {\bm \rho}}

\newcommand{\OHR}{\Omega_{\text{HR}}}
\newcommand{\OLR}{\Omega_{\text{LR}}}

\newcommand{\RR}{\mathbb{R}} 

\newcommand{\cP}{\mathcal{P}} 

\usepackage[pdftex,
  bookmarks=true,
  bookmarksnumbered=true,
  bookmarksopen=true,
  hypertexnames=true,
  breaklinks=true,
  colorlinks=true,
  linkcolor=red,
  citecolor=green,
  pagecolor=blue,
  pagebackref=false,
  urlcolor=red]{hyperref}

 \iccvfinalcopy 

\def\iccvPaperID{0006} 

\begin{document}

\title{Depth Super-Resolution Meets Uncalibrated Photometric Stereo}


\author{Songyou Peng$^{1,2}$ \qquad Bjoern Haefner$^{1}$ \qquad Yvain Qu\'{e}au$^{1}$ \qquad Daniel Cremers$^{1}$
\\
$^{1}$ Computer Vision Group, Technical University of Munich\\
$^{2}$ Erasmus Mundus Masters in Vision and Robotics (VIBOT)\\
{\vspace{-1.2em}\tt\footnotesize \{songyou.peng, bjoern.haefner, yvain.queau, cremers\}@in.tum.de}
}

\twocolumn[{%
\renewcommand\twocolumn[1][]{#1}%
\maketitle

\begin{center}
    \setlength{\tabcolsep}{0.1em} 
    {\renewcommand{\arraystretch}{0.6}
      \begin{tabular}{@{\hspace{-3.0em}}c@{\hspace{0em}}c@{\hspace{0em}}c@{\hspace{0.0em}}}
      \begin{minipage}{0.5\textwidth}
		\centering
		\includegraphics[width = 0.72\linewidth]{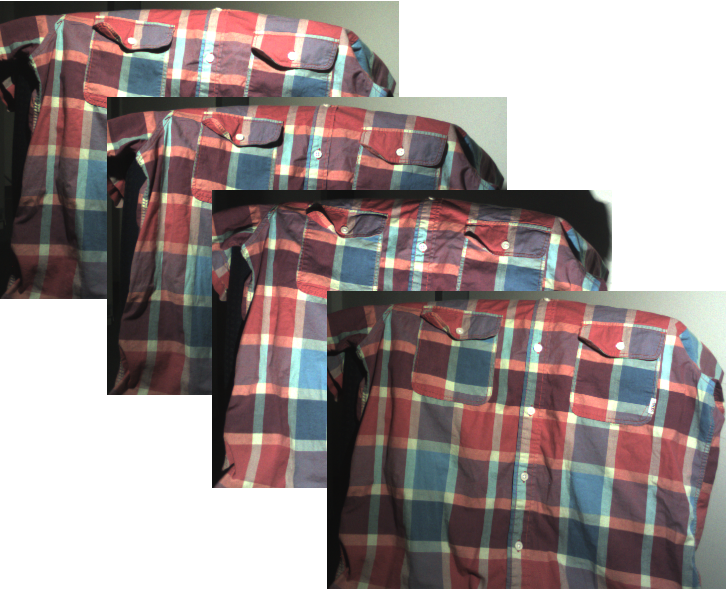}\\
		\includegraphics[width = 0.16\linewidth]{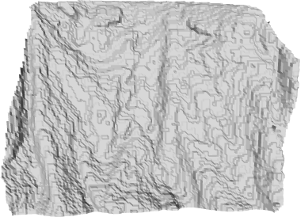}\hspace{0.2em}
		\includegraphics[width = 0.16\linewidth]{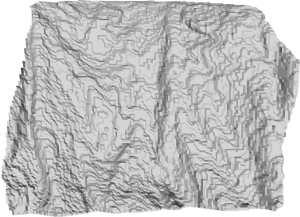}\hspace{0.2em}
		\includegraphics[width = 0.16\linewidth]{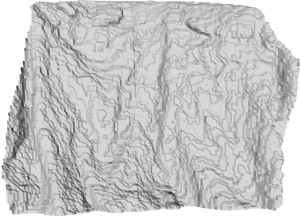}\hspace{0.2em}
		\includegraphics[width = 0.16\linewidth]{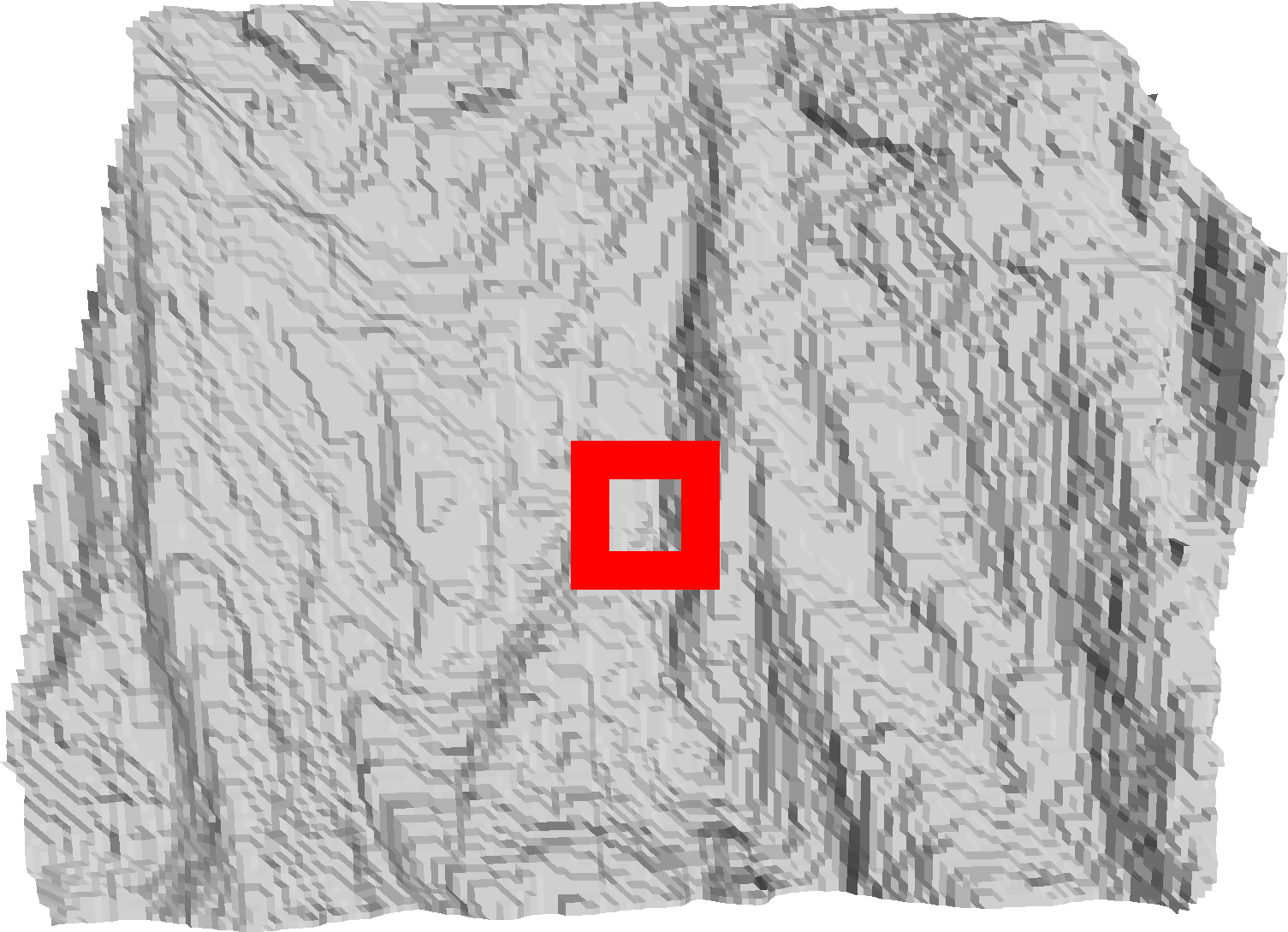}	
	\end{minipage}&
		\begin{minipage}{0.23\textwidth}
		\centering
		\includegraphics[width = 0.84\linewidth]{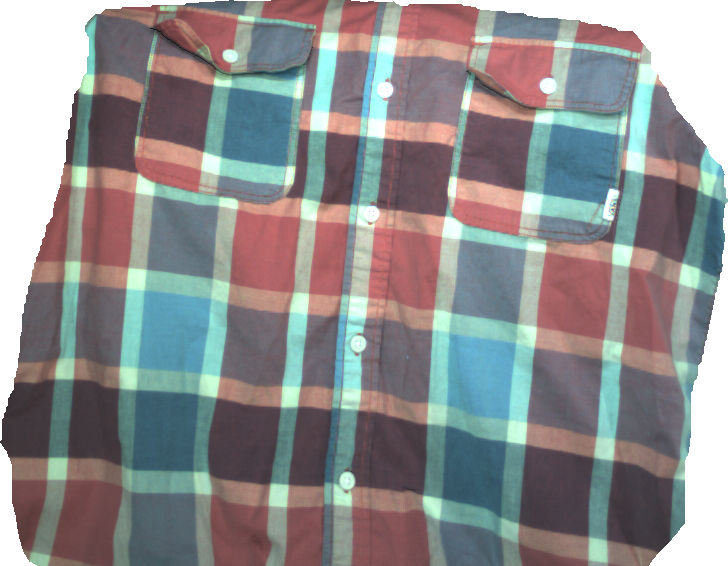}\\
		\includegraphics[width = 0.84\linewidth]{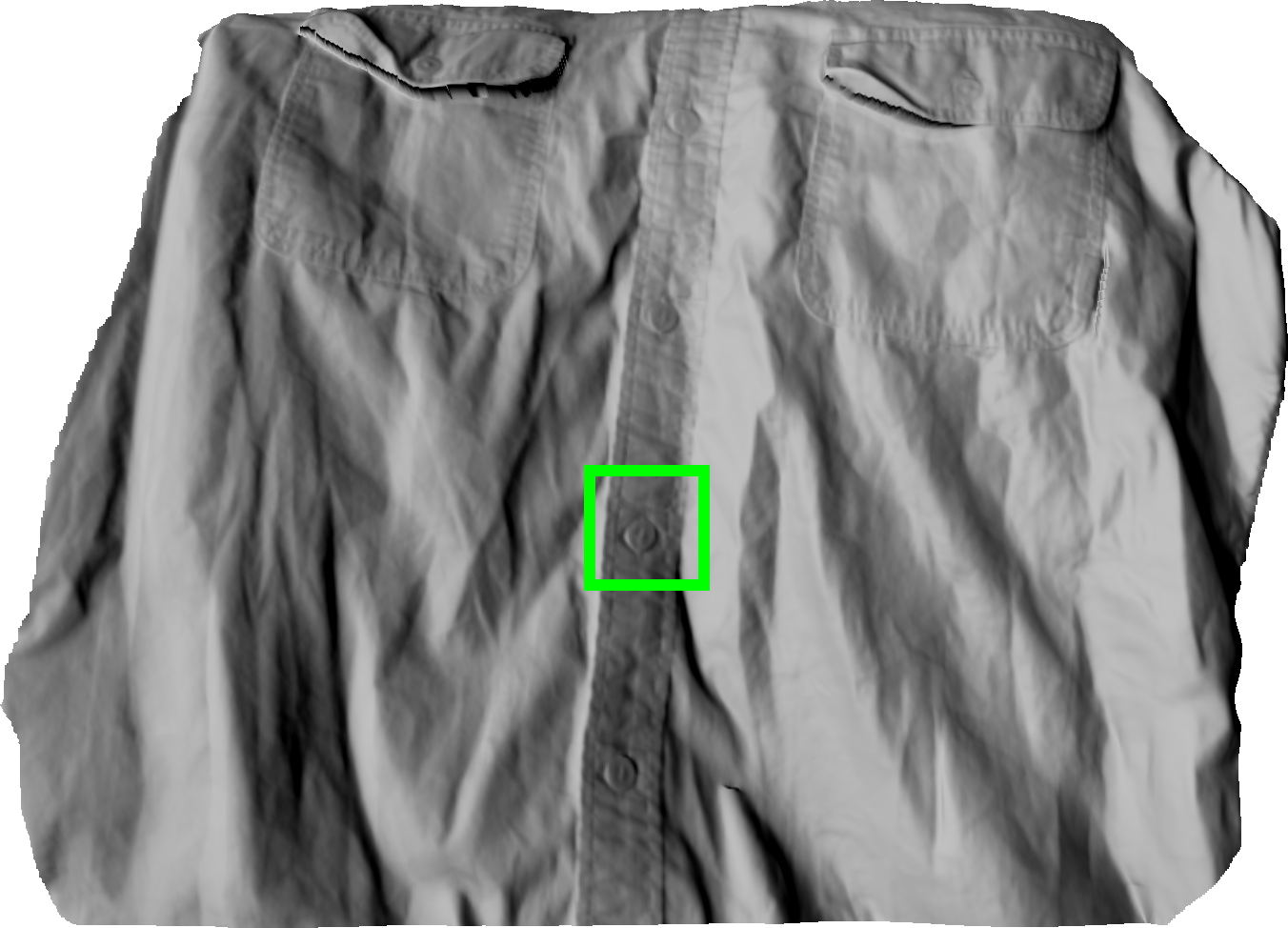}\\
		\includegraphics[width = 0.24\linewidth]{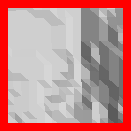}\hspace{0.5em}
		\includegraphics[width = 0.24\linewidth]{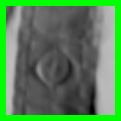}	
	\end{minipage}	&
		\begin{minipage}{0.23\textwidth}
		\centering
		\includegraphics[width = 0.75\linewidth]{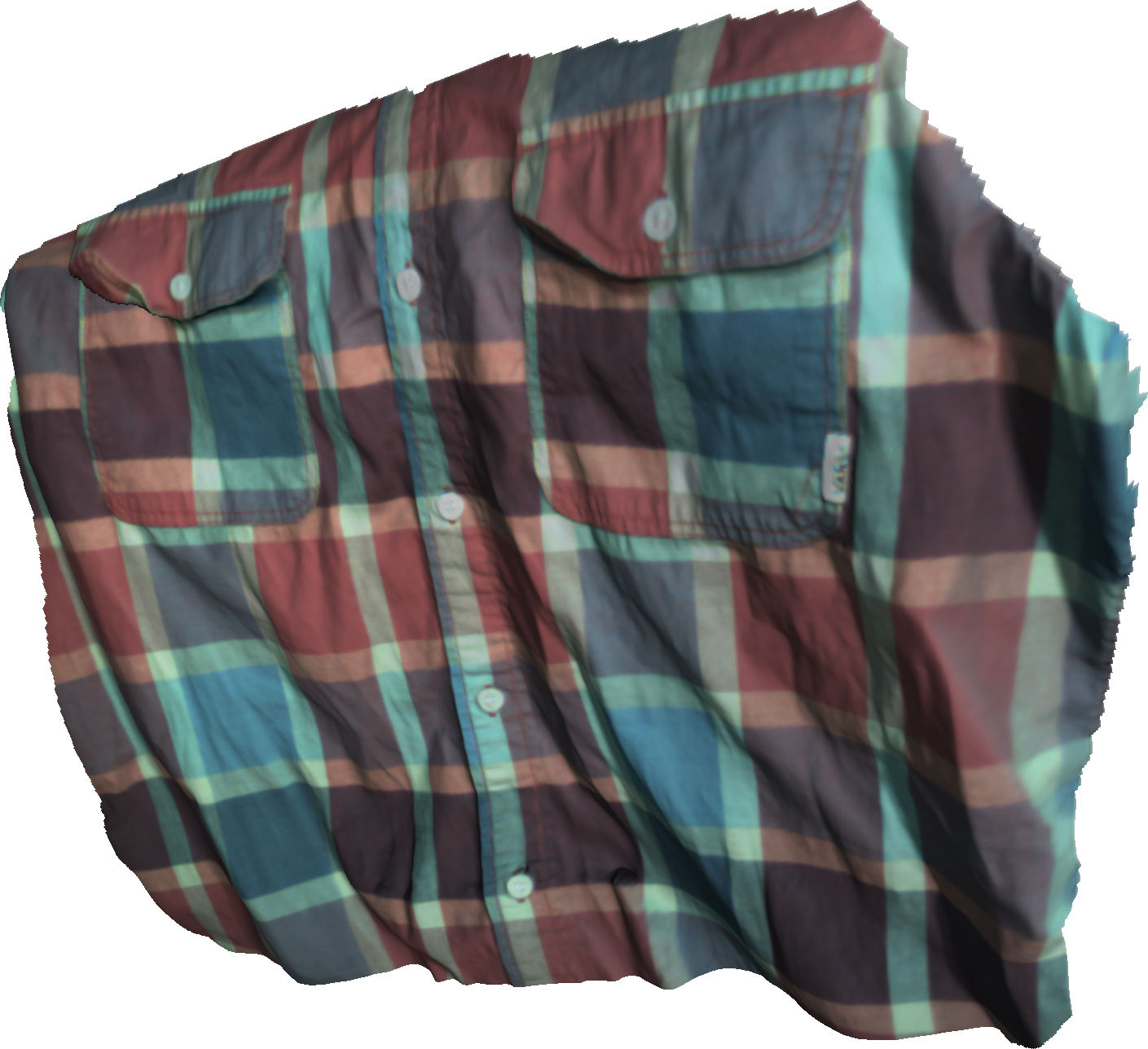}\\
		\includegraphics[width = 0.75\linewidth]{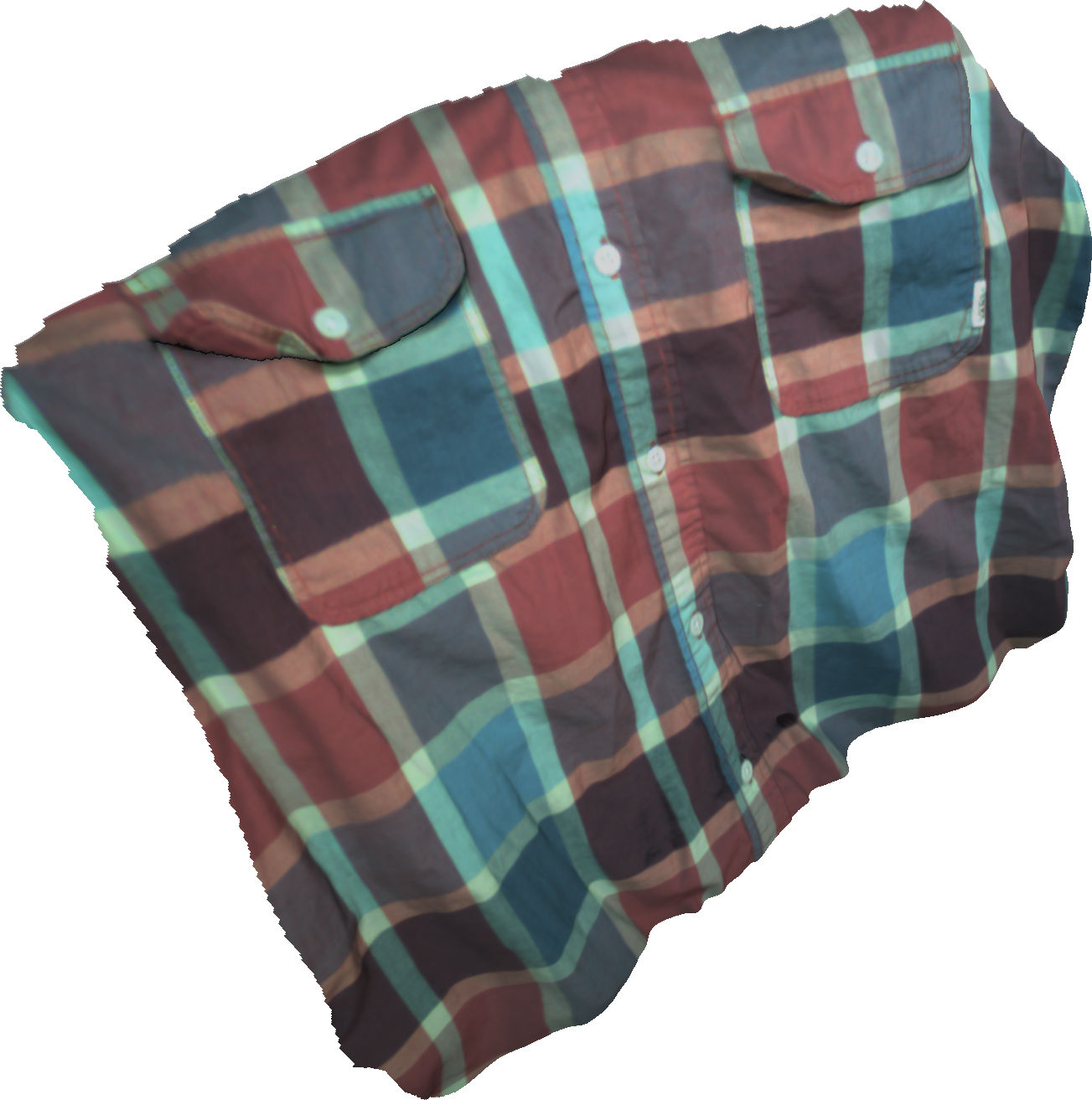}
	\end{minipage}		\\
          
          {Input HR RGB images and LR depth maps} & {Output HR albedo and depth maps} & {Relighting}\\          
          
      \end{tabular}
    }
    \captionof{figure}{Given an RGB-D sequence of $n \geq 4$ low-resolution ($320 \times 240$ $px$) depth maps and high-resolution ($1280 \times 1024$ $px$) RGB images acquired from the same viewing angle but under varying, unknown lighting, high-resolution depth and reflectance maps are estimated by combining super-resolution and photometric stereo within a variational framework.}
    \label{fig:teaser}
\end{center}%
}]

\begin{abstract}
   A novel depth super-resolution approach for RGB-D sensors is presented. It disambiguates depth super-resolution through high-resolution photometric clues and, symmetrically, it disambiguates uncalibrated photometric stereo through low-resolution depth cues. To this end, an {RGB-D} sequence is acquired from the same viewing angle, while illuminating the scene from various uncalibrated directions. This sequence is handled by a variational framework which fits high-resolution shape and reflectance, as well as lighting, to both the low-resolution depth measurements and the high-resolution RGB ones. The key novelty consists in a new PDE-based photometric stereo regularizer which implicitly ensures surface regularity. This allows to carry out depth super-resolution in a purely data-driven manner, without the need for any ad-hoc prior or material calibration. Real-world experiments are carried out using an out-of-the-box RGB-D sensor and a hand-held LED light source.  
\end{abstract}

\section{Introduction}

RGB-D sensors such as Microsoft Kinect or Asus Xtion Pro Live have become a popular way to acquire colored 3D-representations of the world at low-cost. Yet, the accuracy of such representations remains limited by two factors. 

First, the depth channel is prone to quantization and noise, and it has a \emph{coarser resolution} than the RGB one. For instance, the Asus Xtion Pro Live sensor provides QVGA ($320 \times 240$ $px$) or VGA ($640 \times 480$ $px$) depth resolution, while it offers SXGA ($1280 \times 1024$ $px$) RGB resolution. Therefore, the RGB image is often downsampled to match the size of the depth channel. Some information is thus lost and the color may appear blurred or even aliased. Alternatively, the low-resolution (LR) depth map can be upsampled to the size of the high-resolution (HR) RGB image. This problem, known as super-resolution, is however ill-posed. 

Second, the RGB image appears shaded due to ambient illumination. This may cause the 3D-reconstruction to look unrealistic in relighting or augmented reality applications. One would rather use \emph{reflectance} in such applications, and not direct RGB (luminance) measurements. 

This work simultaneously addresses both issues, by appropriately combining depth super-resolution and uncalibrated photometric stereo. It is shown that, by considering an RGB-D sequence acquired from the same viewing angle but under varying, unknown lighting, the LR depth measurements can be super-resolved without resorting to any ad-hoc prior or calibration. Reflectance and lighting are obtained as by-products. This is illustrated in Figure~\ref{fig:teaser}, and formalized as follows.

\vspace*{-1em}
\paragraph{Problem Statement --} Given a set of $n \geq 4$ HR RGB images $\bI^i:\,\OHR \subset \RR^2 \to \mathbb{R}^3,~i \in \{1,\dots,n\}$, and aligned LR depth maps $z_0^i:\,\OLR \subset \OHR \to \mathbb{R},~i \in \{1,\dots,n\}$, acquired from the same viewing angle but under varying, unknown lighting, estimate an HR depth map $z:\,\OHR \to \RR$, an HR reflectance map $\brho:\,\OHR \to \RR^3$, and colored first-order spherical harmonics lighting $\{\mathbf{l}^i \in \RR^{12}\}_i$.

\vspace*{-1em}
\paragraph{Contribution and Organization of the Paper --} After discussing related work in Section~\ref{sec:2}, we propose in Section~\ref{sec:3} the new variational model~\eqref{eq:14} for joint depth super-resolution and uncalibrated photometric stereo. It combines a super-resolution fidelity term with a tailored PDE-based regularization term relying on photometric stereo. While the former ensures consistency between the sought HR depth map and the LR ones,  the latter ensures that the sought HR depth map is both regular and consistent with the HR RGB images. Herein, low-resolution depth clues (resp., high-resolution photometric clues) act as natural disambiguation tools for uncalibrated photometric stereo (resp., depth super-resolution). This variational approach is evaluated in Section~\ref{sec:4} against challenging synthetic and real-world datasets. Eventually, Section~\ref{sec:5} summarizes our achievements and suggests future research directions.

\section{Related Work}
\label{sec:2}


\paragraph{Depth Super-resolution --}

The most common way to achieve super-resolution consists of acquiring $n$ LR measurements, and combine them into a single HR one. Starting from the seminal work of Tsai and Huang using Fourier analysis~\cite{Tsai1984}, various mathematical tools have been proposed for this task \cite{ouwerkerk2006}. In the present work, we follow the variational approach. 

The LR measurements $\{z_0^i\}_{i \in \{1,\dots,n\}}$ are  assumed to result from downsampling and convolving an HR signal $z$, up to an additive, zero-mean homoskedastic Gaussian noise with standard deviation $\sigma_z$: 
\begin{equation}
z_0^i = K z + \varepsilon_z^i,~\forall i \in \{1,\dots,m\},  
\label{eq:1}
\end{equation}
where $K$ is the downsampling $/$ convolution kernel, and $\varepsilon_z^i(\bp) \sim \mathcal{N}(0,{\sigma_z}^2),\, \bp \in \OHR$. In the present work, $Kz$ can be described for each low-resolution pixel as a weighted sum over the corresponding super-resolution pixels, see~\cite{unger2010} for a detailed explanation.

Estimating the HR signal $z$ comes down to solving the inverse problem~\eqref{eq:1}, which is ill-posed. A standard way to ensure well-posedness consists in introducing a prior on the HR signal and resorting to Bayesian inference. Such a strategy yields a variational problem of the form:
\begin{equation}
\underset{z:\, \OHR \to \RR}{\min}~ \mathcal{R}(z) + \dfrac{1}{2n} \sum_{i=1}^n \| Kz - z_0^i \|^2_{\ell^2(\OLR)}, 
\label{eq:2}
\end{equation}
where $\mathcal{R}$ is a regularization term and $\| \cdotp \|^2_{\ell^2(\OLR)}$ is the $\ell^2$-norm over the LR domain $\OLR$. A typical choice for the regularizer is the total variation (TV) $\mathcal{R}(z) = \lambda \| \nabla z \|_{\ell^1(\OHR)}$~\cite{Marquina2008}, with $\lambda >0$ a tuning parameter and $\nabla$ the gradient operator. This is essentially equivalent to assuming that the solution is  piecewise-constant. 

Super-resolution techniques have found numerous applications ranging from surveillance~\cite{Cristani2004} to medical imaging~\cite{Greenspan2008}, remote sensing~\cite{Fablet2015} or, closer to our proposal, 3D-reconstruction using multi-view stereo~\cite{Goldlucke2014} and RGB-D sensors~\cite{Maier2015}. In such applications where HR RGB measurements $\{\bI^i\}_{i \in \{1,\dots,n\}}$ are available, they may be used as ``guides'' for depth super-resolution. For instance, the following anisotropic RGB image driven Huber-loss regularization term is advocated in~\cite{werlberger2009}:
\begin{equation}
\mathcal{R}(z) = \int_{\Omega_{\text{HR}}} \!\!\!\!\!\! H_\varepsilon(z)
\mathrm{d}\mathbf{p},~ H_\varepsilon(z):= \begin{cases} \frac{\left\lVert
D\nabla z \right\rVert^2}{2\varepsilon} & \!\!\!\!\!\!\!\!\!\!\!\! \text{if
$\left\lVert D\nabla z \right\rVert\leq\varepsilon$} \\ \left\lVert D\nabla
z \right\rVert-\frac\varepsilon2 & \text{else, }\end{cases},
\label{eq:3}
\end{equation}
where $D = \exp\left(\alpha \lVert \nabla\bar\bI \rVert^\beta \right)
\bv\bv^t + \bv^\perp\left(\bv^\perp\right)^t$ with
$\bv=\frac{\nabla\bar\bI}{\lVert\nabla\bar\bI\rVert}$, $\bv^\perp$ a normal
vector to $\bv$, $\bar\bI = \text{mean}\left(\{\bI^i\}_{i \in
\{1,\dots,n\}}\right)$, $(\alpha,\beta)$ some parameters and $\lVert \cdot
\rVert$ is the standard (Euclidean) norm.x
This regularizer tends to smooth $z$ along, but not across edges and corners of the corresponding RGB image. Other image-based regularizers are also discussed in~\cite{Park2014}. Employing the RGB measurements, which have a built-in higher resolution than the depth ones, indeed seems natural. However, this is not straightforward because image variations not only reflect shape variations, but also the interactions between light and matter. This is where photometric techniques come into play. 

\vspace*{-1em}
\paragraph{Uncalibrated Photometric Stereo --}

Inferring shape solely from image clues is an ill-posed problem, known as shape-from-shading~\cite{Horn1989}. It is impossible to unambiguously estimate shape from a single image, even when the reflectance is known. A natural way to disambiguate shape-from-shading is to consider not just one, but multiple images, obtained under varying lighting. This method is known as photometric stereo~\cite{Woodham1980}. Assuming Lambertian reflectance with only additive, zero-mean homoskedastic Gaussian noise with standard deviation $\sigma_I$ (no specularity or cast-shadow), and approximating lighting by first-order spherical harmonics~\cite{Basri2007}, the following image formation model can be considered:
\begin{equation}
I^i_\star(\bp) = \rho_\star(\bp) \, \mathbf{l}^i_\star \cdot \begin{bmatrix} \mathbf{n}(\bp) \\ 1 \end{bmatrix} + \varepsilon_\star^i(\bp),
\label{eq:4}
\end{equation}
with $(i,\star,\bp) \in \{1,\dots,n\} \times \{R,G,B\} \times \OHR$ the indices of the images, channel and pixel, $I^i_\star(\bp) \in \RR$ the $i$-th image value in channel $\star$ at pixel $\bp$, $\rho_\star:\,\OHR \to \RR$ the albedo (Lambertian reflectance) map in channel $\star$, $\bl^i_\star \in \RR^4$ the $i$-th lighting vector in channel $\star$, $\bn(\bp) \in \mathbb{S}^2$ the unit-length outward normal at the surface point conjugate to pixel $\bp$, and $\varepsilon_\star^i(\bp) \sim \mathcal{N}(0,{\sigma_I}^2)$. 

Uncalibrated photometric stereo aims at inferring reflectance, shape and lighting from the images, by solving the system of equations~\eqref{eq:4}. Unfortunately, this problem is ill-posed: it can be solved only up to a linear ambiguity~\cite{Hayakawa1994}. It is common to further enforce surface regularity~\cite{Yuille1997}, which reduces the ambiguity to a generalized bas-relief (GBR) one under directional lighting~\cite{Belhumeur1999}, and to a Lorenz one under spherical harmonics lighting~\cite{Basri2007}. Resolution of such ambiguities by resorting to additional priors~\cite{Alldrin2007,Papadhimitri2014b}, and extensions to non-Lambertian reflectances~\cite{Lu2017}, remain active research topics. It has also been shown recently in~\cite{Queau2017} that PDE-based approaches may be worthwhile for uncalibrated photometric stereo, because they implicitly enforce integrability and thus naturally reduce ambiguities. 

\vspace*{-1em}
\paragraph{Photometric RGB-D Sensing --} Depth sensing improvement by shading analysis has been tackled in many recent works~\cite{Han2013,Kim2015,Or-el2015,Wu2014,Yu2013}. However, such methods do not actively control lighting, and thus they suffer from the same ambiguity as shape-from-shading. In particular, a smoothness prior on reflectance is always required. We will see in Section~\ref{sec:4} that this considerably limits applicability. To unambiguously estimate reflectance using an RGB-D sensor, there is no other choice but to actively control lighting \ie, to resort to photometric stereo~\cite{Anderson2011,Chatterjee2015}.

\vspace*{-1em}
\paragraph{Photometric Super-Resolution --}
Super-resolution and photometric stereo have been widely studied, but rarely together. Some authors super-resolve the photometric stereo results~\cite{Tan2008}, and others generate HR images using photometric stereo~\cite{Chaudhuri2005}, but few employ LR depth clues. The only work in that direction is that in~\cite{Lu2013}, where calibrated photometric stereo and structured light sensing are combined. However, this involves a non-standard setup and careful lighting calibration, and reflectance is assumed to be uniform. In contrast, we provide in the next section working tools for out-of-the-box RGB-D sensors and surfaces with spatially-varying reflectance. Therein, it is only assumed that the LR depth maps are aligned with the HR RGB images and that the RGB sensor's intrinsics are known (both, the warping function and the intrinsics can be accessed \eg, using OpenNI 2 for ROS).

\section{A Variational Framework for Photometric Stereo-Aware Depth Super-resolution}
\label{sec:3}


The main contribution of this work is now presented. It consists of the variational approach~\eqref{eq:14} to joint depth super-resolution and uncalibrated photometric stereo. This variational framework involves a regularizer built upon the PDE-based photometric stereo model described hereafter.

\subsection{PDE-based Photometric Stereo with First-Order Spherical Harmonics Lighting}
\label{sec:3.1}

The super-resolution fidelity term in~\eqref{eq:2} is expressed in terms of depth, instead of normals. Therefore, we resort to a differential photometric stereo approach to design the regularization term. Let us first show how to express~\eqref{eq:4} as a system of nonlinear PDEs in ${z}$, $\brho$ and $\{\bl^i\}_{i \in \{1,\dots,n\}}$ over $\OHR$ which have the following form:
\begin{equation}
\bA^i({z},\brho,\bl^i)^\top \begin{bmatrix} \nabla {z} \\ {z} \end{bmatrix} \\  = \bb^i( \brho,\bl^i)  + {\bm \varepsilon}^i,\,i \in \{1,\dots,n\}, 
\label{eq:5}
\end{equation}
where $\bA^i({z},\brho,\bl^i):\OHR \to \RR^{3 \times 3}$ and $\bb^i(\brho,\bl^i):\OHR \to \RR^{3}$ are fields which depend on the unknowns, and each ${\bm \varepsilon}^i,\,i \in \{1,\dots,n\}$ is a random $\OHR \to \RR^3$ homoskedastic Gaussian vector field with zero-mean and diagonal covariance matrix $\text{Diag}( {\sigma_I}^2,{\sigma_I}^2,{\sigma_I}^2)$.

Under perspective projection, the normal in~\eqref{eq:4} reads:
\begin{equation}
\bn(\bp) = \frac{1}{d(z)(\bp)} \begin{bmatrix}
f \nabla z(\bp) \\
-z(\bp) - \nabla z(\bp) \cdotp \left( \bp - \bp^0 \right) 
\end{bmatrix},
\label{eq:6}
\end{equation}
with $f>0$ the focal length, $\bp^0 \in \OHR$ the principal point, and where $d({z})(\bp)$ is equal to the norm of the bracket (unit-length constraint). Plugging~\eqref{eq:6} into~\eqref{eq:4}, the nonlinear system of PDEs~\eqref{eq:5} is obtained, with, $\forall \bp \in \OHR$:
\begin{align}
\bA^i({z},\brho,\bl^i)(\bp)  & =  
\dfrac{1}{d(z)(\bp)} 
\Bigg(
f\! \begin{bmatrix} l^i_{R,1} & l^i_{G,1} & l^i_{B,1} \\ l^i_{R,2} & l^i_{G,2} & l^i_{B,2} \\ 0 & 0 & 0 \end{bmatrix} \nonumber \\ 
 & \!\!\!\!\!\!\!\!\!\!\!\!\!\!\!\! - \begin{bmatrix}\bp\!-\!\bp^0 \\ 1 \end{bmatrix} \left[l^i_{R,3},\,l^i_{G,3},\,l^i_{B,3}\right] \! \Bigg) 
\text{Diag}(\brho(\bp)),
\label{eq:7} \\
\bb^i(\brho,\bl^i)(\bp) & =  \bI^i(\bp) \!-\! \begin{bmatrix} l^i_{R,4} & & \\ & l^i_{G,4} & \\ & & l^i_{B,4} \end{bmatrix} \, \brho(\bp),
\label{eq:8}
\end{align} 
where $\bI^i(\bp) = \left[I^i_R(\bp),I^i_G(\bp),I^i_B(\bp)\right]^\top \in \RR^3$, $\brho(\bp) = \left[\rho_R(\bp),\rho_G(\bp),\rho_B(\bp)\right]^\top \in \RR^3$, and $\bl^i = \left[ \left[\bl^i_R\right]^\top,\left[\bl^i_G\right]^\top,\left[\bl^i_B\right]^\top \right]^\top \in \RR^{12}$.

Let us remark that Model~\eqref{eq:5} is slightly more complex than previous PDE-based photometric stereo models such as the one in~\cite{Queau2017}, because we consider first-order spherical harmonics lighting. In practice, this allows us to cope with much less restricted environments, for instance in the presence of strong ambient lighting.

\subsection{Proposed Variational Framework}
\label{sec:3.2}




For the numerical solution, we follow a purely data-driven (maximum likelihood) variational approach. By independence of image and depth measurements, and of reflectance and lighting, the likelihood factorizes as follows:
\begin{equation}
\cP(\{z_0^i,\bI\}_i \vert z,\brho,\{\bl^i\}_i ) = \cP(\{\bI\}_i \vert z,\brho,\{\bl^i\}_i) ~ \cP(\{z_0^i\}_i \vert z). 
\label{eq:10}
\end{equation}

In addition, Equations~\eqref{eq:1} and~\eqref{eq:5} induce:
\begin{align}
& \cP(\{\bI\}_i \vert z,\brho,\{\bl^i\}_i) 
  = \left( 2 \pi {\sigma_I}^2 \right)^{-\frac{3 n |\OHR|}{2}}
   \exp \bigg\{ \left(-2 {\sigma_I}^2 \right)^{-1} \nonumber \\[-.5em]
& \quad  \sum_{i=1}^n \left\| \bA^i(z,\brho,\bl^i)^\top \begin{bmatrix} \nabla z \\ z \end{bmatrix} - \bb^i(\brho,\bl^i) \right\|_{\ell^2(\OHR)} \bigg\}, \label{eq:11}  \\
& \cP(\{z_0^i\}_i \vert z) =  \left(2 \pi {\sigma_z}^2\right)^{-\frac{n |\OLR|}{2}} \exp \bigg\{  \left(2 {\sigma_z}^2 \right)^{-1} \nonumber \\[-1em]
& \quad  \sum_{i=1}^n \left\| Kz - z_0^i \right\|_{\ell^2(\OLR)} \bigg\},  
\label{eq:12} 
\end{align}
where $|\cdotp|$ denotes cardinality. By further denoting: 
\begin{equation}
  \lambda = \frac{{\sigma_z}^2}{{\sigma_I}^2},
  \label{eq:13}
\end{equation}
and since maximizing likelihood~\eqref{eq:10} is equivalent to minimizing its negative logarithm, we obtain from Equations~\eqref{eq:10} to~\eqref{eq:13} the following variational model for joint depth super-resolution, reflectance and lighting estimation:
\begin{align}
\underset{\substack{z:\, \OHR \to \RR \\ \brho:\, \OHR \to \RR^3 \\ \{ \bl^i \in \RR^{12}\}_i }}{\min}\Bigg\{ & ~ \lambda \displaystyle\sum_{i=1}^n  \left\| \bA^i({z},\brho,\bl^i)^\top \begin{bmatrix} \nabla {z} \\ {z} \end{bmatrix} - \bb^i( \brho,\bl^i)\right\|^2_{\ell^2(\OHR)} \nonumber \\[-2em]
 & + \displaystyle\sum_{i=1}^n \| Kz - z_0^i \|^2_{\ell^2(\OLR)} \Bigg\}. 
\label{eq:14}
\end{align}

Problem~\eqref{eq:14} yields the ill-posed uncalibrated photometric stereo one if $\lambda = +\infty$, and the ill-posed super-resolution one if $\lambda = 0$. We conjecture that any choice in between disambiguates both problems, but proving these conjectures is beyond the scope of this proof of concept work. 

\subsection{Alternating Optimization Strategy}
\label{sec:3.3}

\begin{figure}[!ht]
  \includegraphics[width = \linewidth]{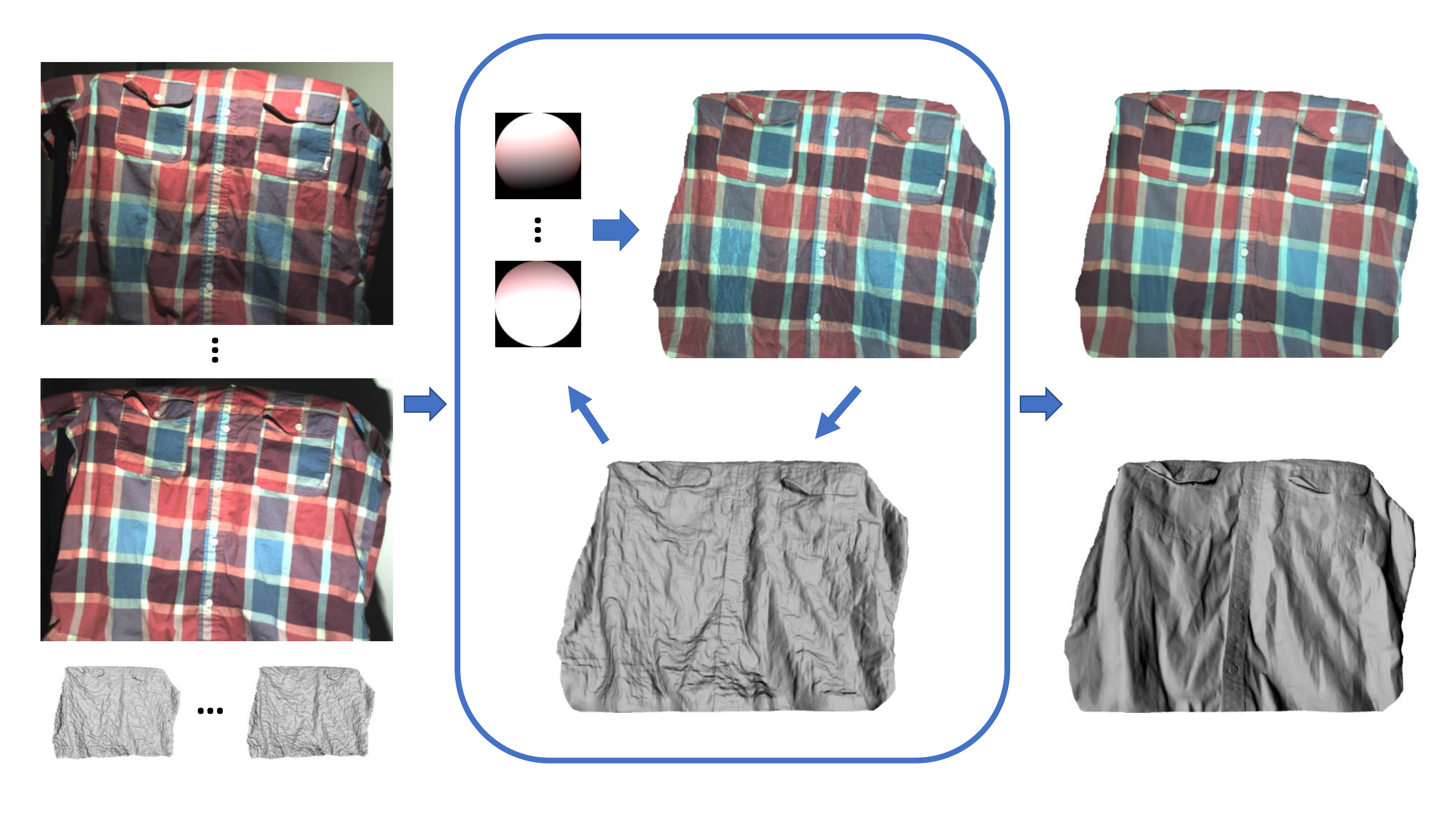}
\caption{Sketch of our optimization framework for HR reflectance and depth estimation, given a series of HR RGB images and LR depth maps. Lighting, HR reflectance and HR depth are sequentially optimized until convergence.}
\label{fig:sketch}
\end{figure}

The variational problem~\eqref{eq:14} is solved iteratively in terms of lighting, reflectance and depth, as illustrated in Figure~\ref{fig:sketch}. Reflectance and lighting updates come down to simple linear least-squares problems. During each depth update, we ``freeze'' the matrix fields $\bA^i$ and $\bl^{i}$ to their current values to obtain a linear (weighted) least-squares problem which is solved using conjugate gradient iterations. Initially, the reflectance is assumed uniformly white ($\brho \equiv 1$) and the depth is obtained by meaning the LR measurements, filling missing values by biharmonic inpainting and eventually upsampling using bicubic interpolation. No initial lighting estimate is required, and the algorithm stops when the relative difference between two successive energy values falls below a threshold set to $0.01$. 

\section{Empirical Validation}
\label{sec:4}


\subsection{Quantitative Evaluation on Synthetic Datasets}

\begin{figure}[!ht]
\begin{minipage}{.7\linewidth}
\begin{tabular}{cc}  
  \includegraphics[width = 0.6\linewidth, trim = 6em 10em 7em 20em, clip]{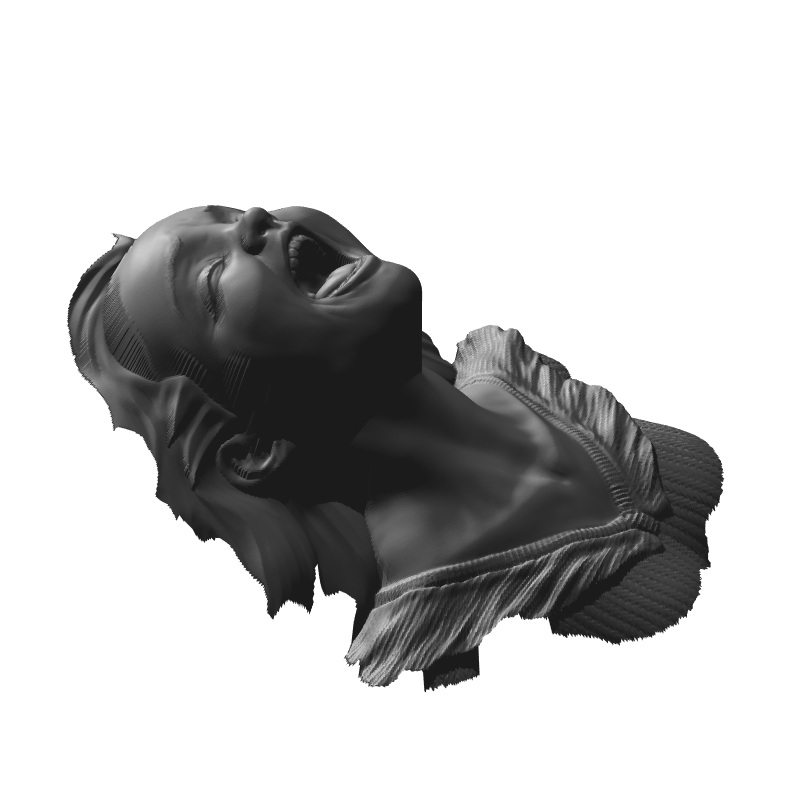} & 
  \includegraphics[width = 0.38\linewidth, trim = 0em 2em 0em 2em, clip, frame]{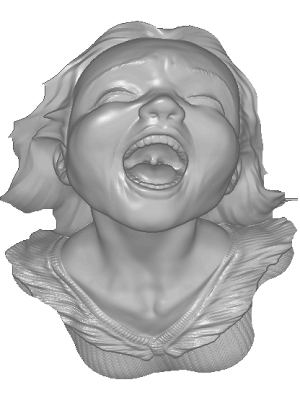} \\
  {\small (a)} &   {\small (b)} 
\end{tabular}
\end{minipage}\qquad
\begin{minipage}{.14\linewidth}
\begin{tabular}{cc}  
  \includegraphics[width = \linewidth, trim = 0.5em 2em 0.5em 2em, clip, frame]{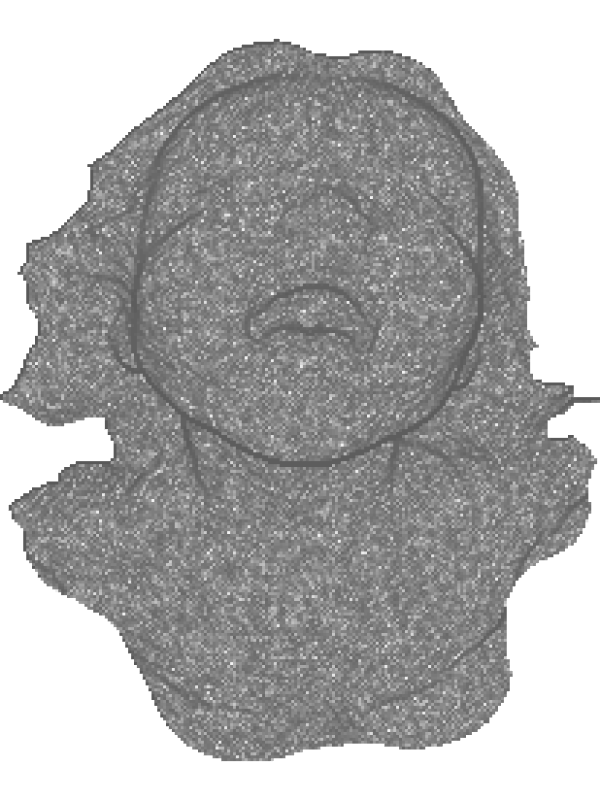} \\
  \includegraphics[width = 0.5\linewidth, trim = 0.25em 1em 0.25em 1em, clip, frame]{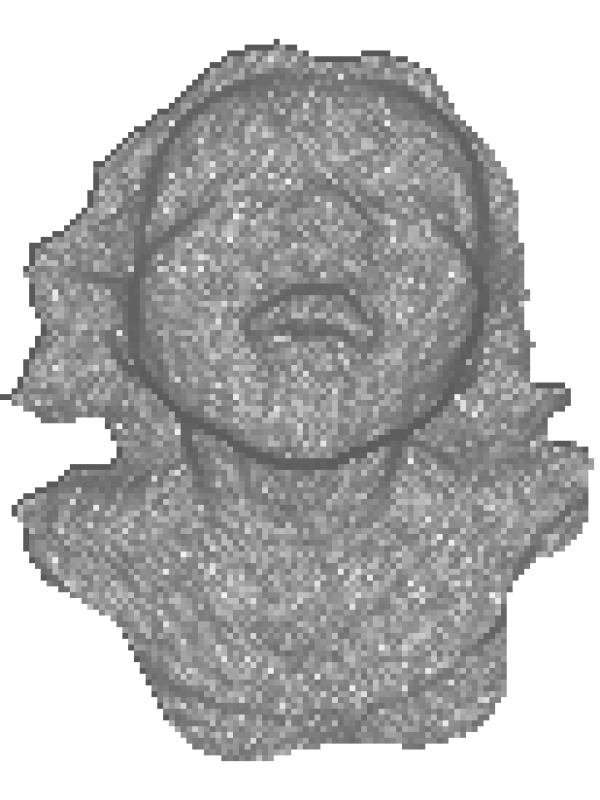} \\
  ~\\
  {\small (c)} 
\end{tabular}
\end{minipage}
\begin{tabular}{@{\hspace{-0.05em}}cc}
  \includegraphics[width = 0.225\linewidth,frame]{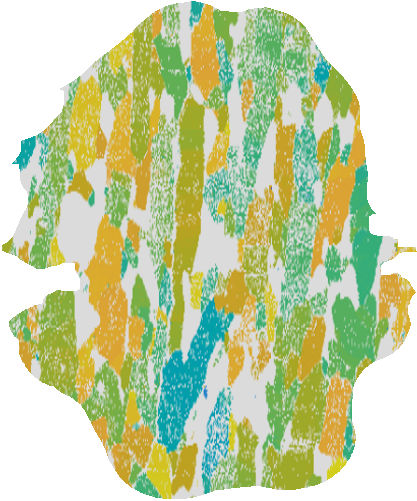}  &
  \includegraphics[width = 0.225\linewidth,frame]{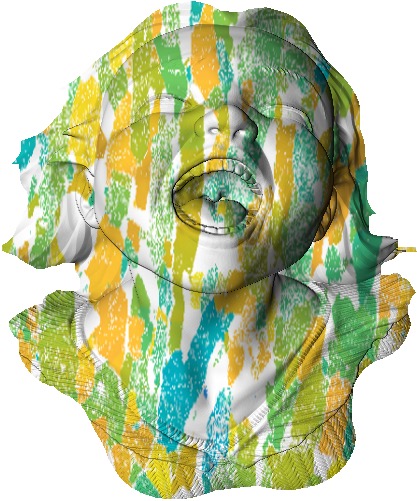}
  \includegraphics[width = 0.225\linewidth,frame]{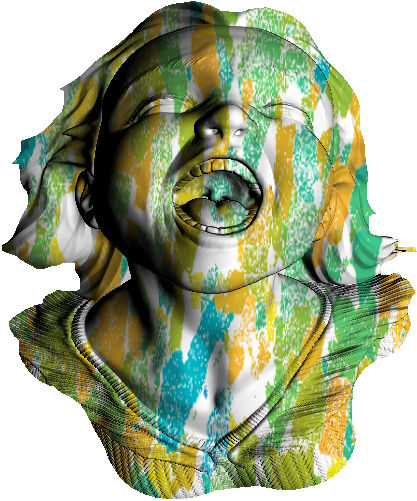}
  \includegraphics[width = 0.225\linewidth,frame]{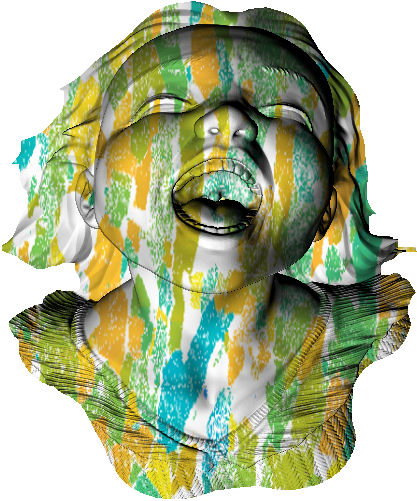} \\      
  {\small (d)} &   {\small (e)}
\end{tabular}
\caption{Synthetic dataset used in our quantitative experiments. (a) 3D-shape. (b) Ground truth HR ($640 \times 480$ $px$) depth map. (c) LR noisy depth maps, for scaling factors of 2 ($320 \times 240$ $px$) and 4 ($160 \times 120$ $px$). (d) HR albedo map (source: {\footnotesize \url{https://mtex-toolbox.github.io/files/doc/EBSDSpatialPlots.html}}). (e) HR photometric stereo images.}
\label{fig:2}
\end{figure}

The public domain ``Joyful Yell'' 3D-shape is first considered. Depth maps with different scale factors are rendered, and noise (Gaussian, zero-mean, with standard deviation $\sigma_z = \alpha_z \|z\|_\infty$, $\alpha_z >0$) is added to each LR depth map. The accuracy of the 3D-reconstruction is evaluated by comparing the HR 3D-reconstruction against the ground-truth. To create the RGB images, we proceed as follows. Using the ground truth depth map, normals are computed by finite differences. Then, random first-order spherical harmonics lighting vectors are generated, and an HR RGB image is taken as ground-truth albedo. All of these are eventually combined into an image generated according to Equation~\eqref{eq:4}. Figure~\ref{fig:2} summarizes this process. 

\begin{figure}[!ht]
\begin{tabular}{cc}
 \!\!\!\!\! \includegraphics[width = 0.48\linewidth, trim = 0em 0em 3.8em 1em, clip]{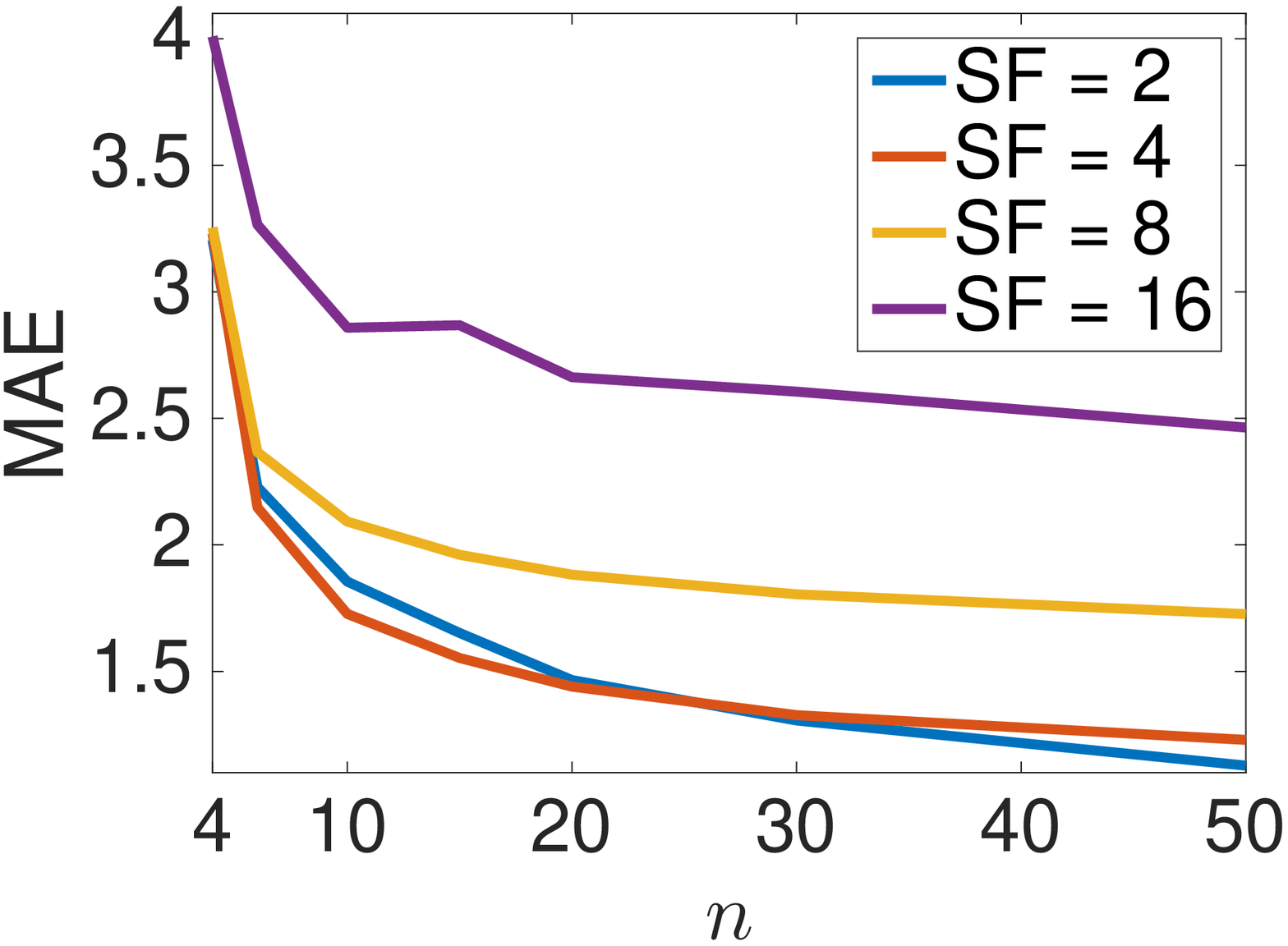} & \!\!\!\!\!\!\!
  \includegraphics[width = 0.48\linewidth, trim = 0em 0em 3.8em 1em, clip]{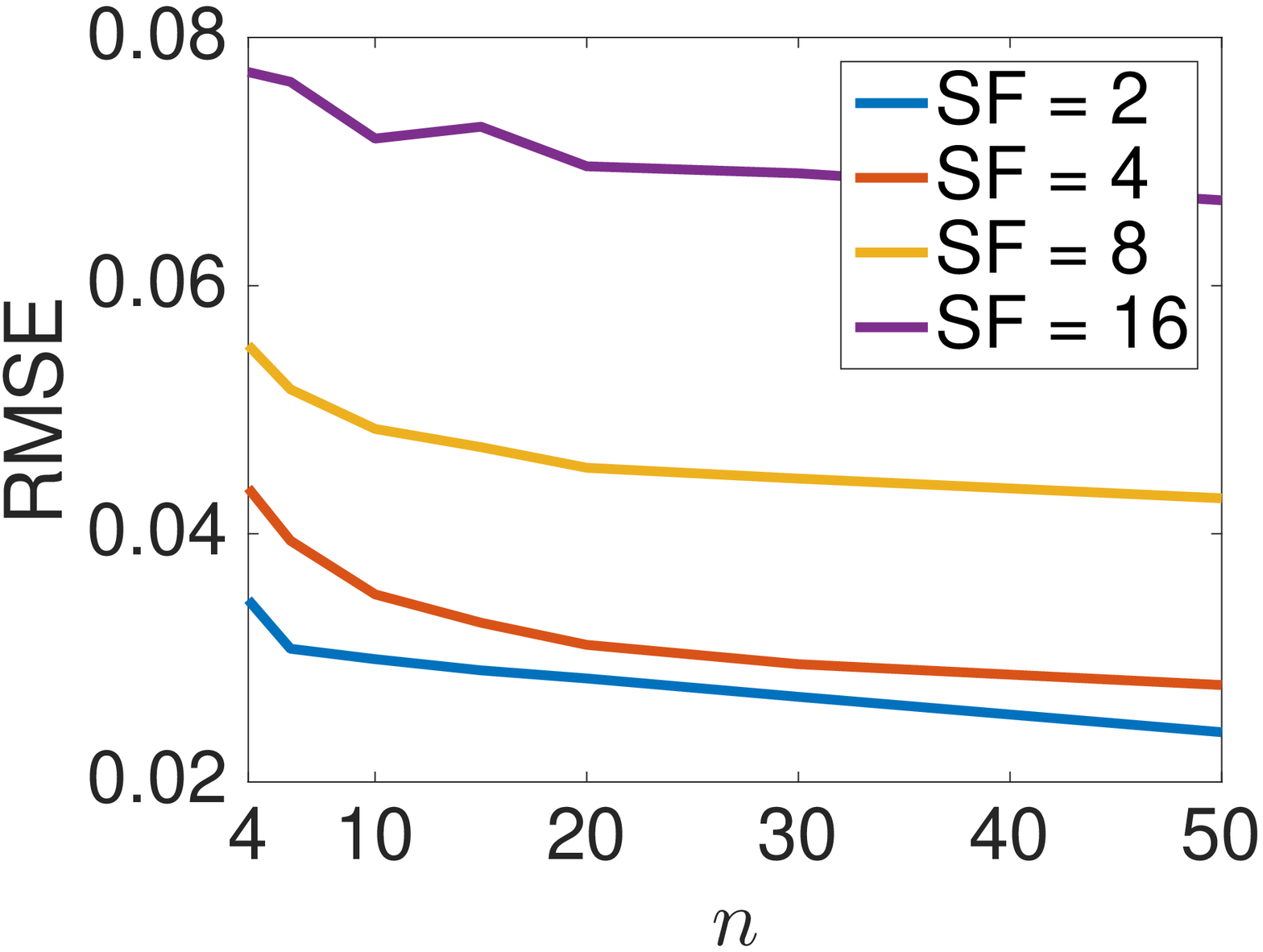}  \\
 \!\!\!\!\! {\small (a)} & \!\!\!\!\!\!\!  {\small (b)} \\
 & \\
   \multicolumn{2}{c}{\includegraphics[width = 0.52\linewidth, trim = 0em 0em 3.8em 1em, clip]{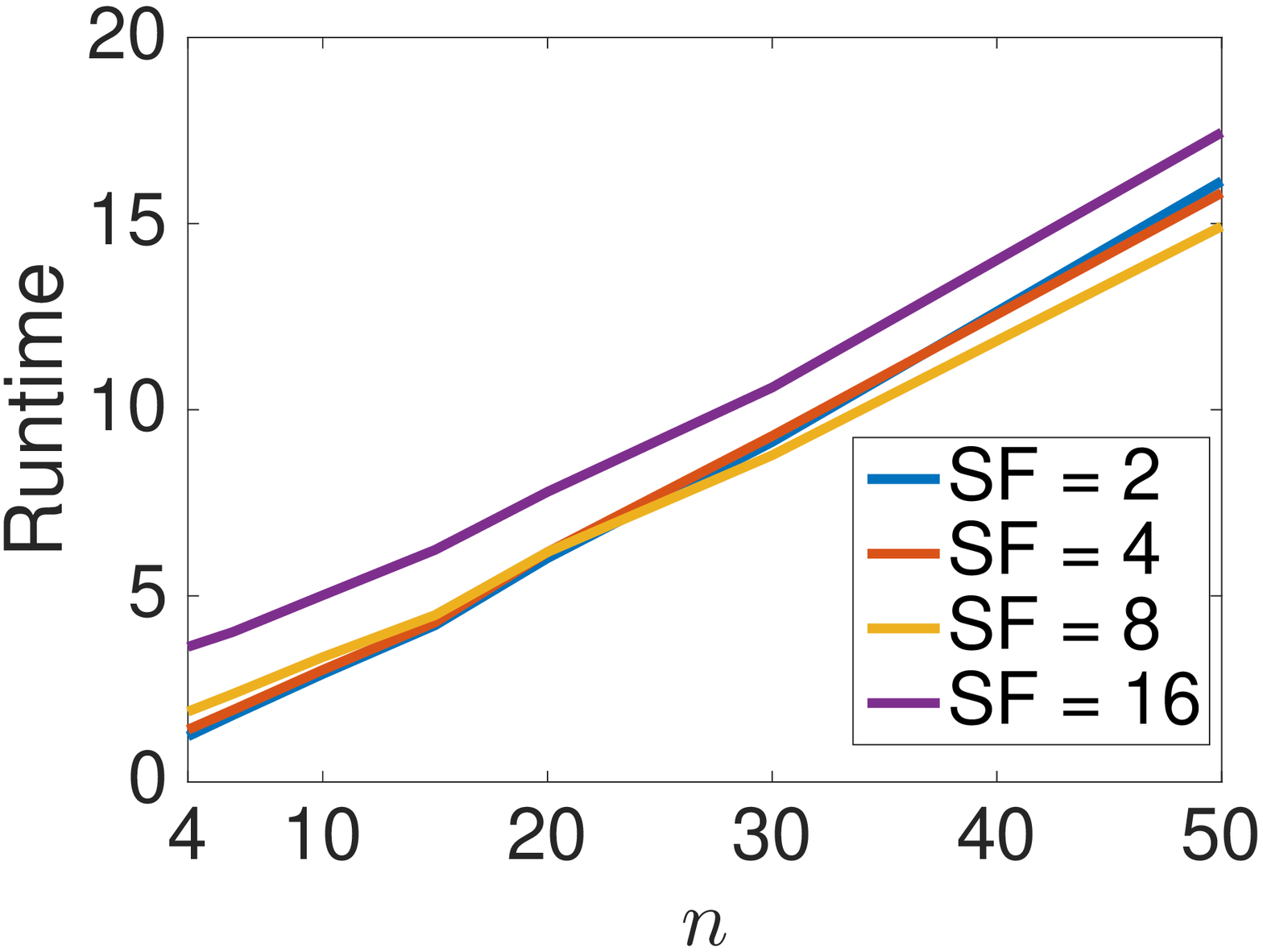}}\\
      \multicolumn{2}{c}{\small (c)}\\ 
\end{tabular}
\caption{Impact of the number $n$ of images on accuracy and computation time, for different scaling factors (SF). (a) Root Mean Square Error (RMSE, in arbitrary units) on depth. (b) Mean Angular Error (MAE, in degrees) on normals. 10 to 30 images are enough to obtain accurate results. (c) Runtime (in seconds) per each iteration as a function of the number $n$ of images.}
\label{fig:3}
\end{figure}

\paragraph{Number of Images --} We first evaluate in Figure~\ref{fig:3} the impact of the number $n$ of depth maps and photometric stereo images on the accuracy of the 3D-reconstruction. Quite obviously, the higher $n$, the more accurate the 3D-reconstruction. However, the runtime (evaluated on a Xeon processor at $3.50$ $GHz$ with $32$ $GB$ of RAM) of each iteration (convergence is reached within at most 15 iterations in all the experiments) increases linearly with $n$. Overall, the choice $n \in [10,30]$  represents a good compromise between accuracy and speed. Besides, somewhat similar results are obtained with a scaling factor of 2 and~4, and only from a scaling factor of 8 the results start to significantly deteriorate. We believe this is not a problem because real-world RGB-D sensors such as the Asus Xtion Pro Live only provide depth maps with resolution $\frac12$ or $\frac14$ that of the HR RGB image.

\begin{figure}[!ht]
  \begin{tabular}{cc}
   \!\!\!\!   \includegraphics[width = 0.48\linewidth, trim = 0em 0em 3em 1em, clip]{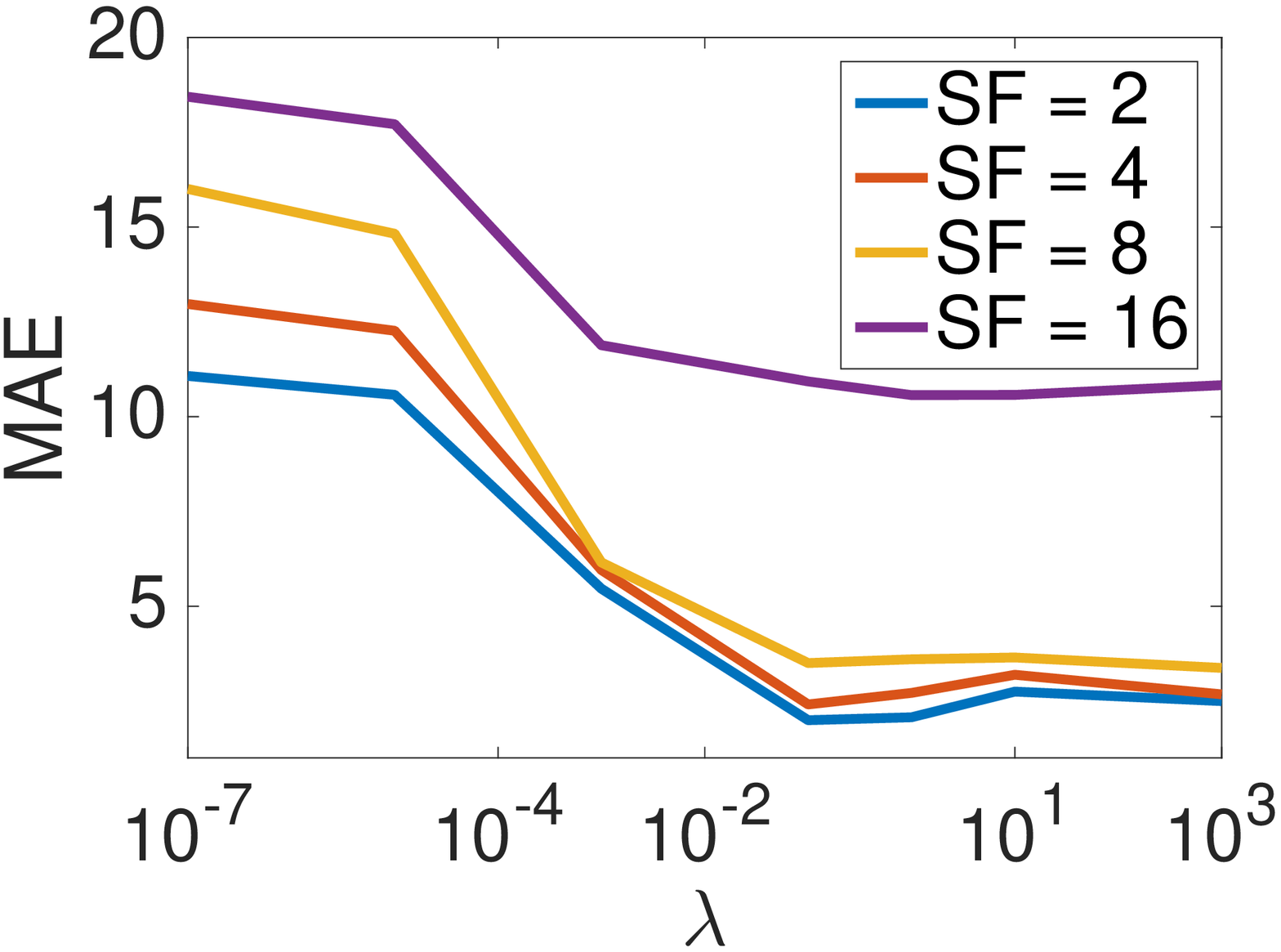}  & \!\!\!\! 
    \includegraphics[width = 0.48\linewidth, trim = 0em 0em 3em 1em, clip]{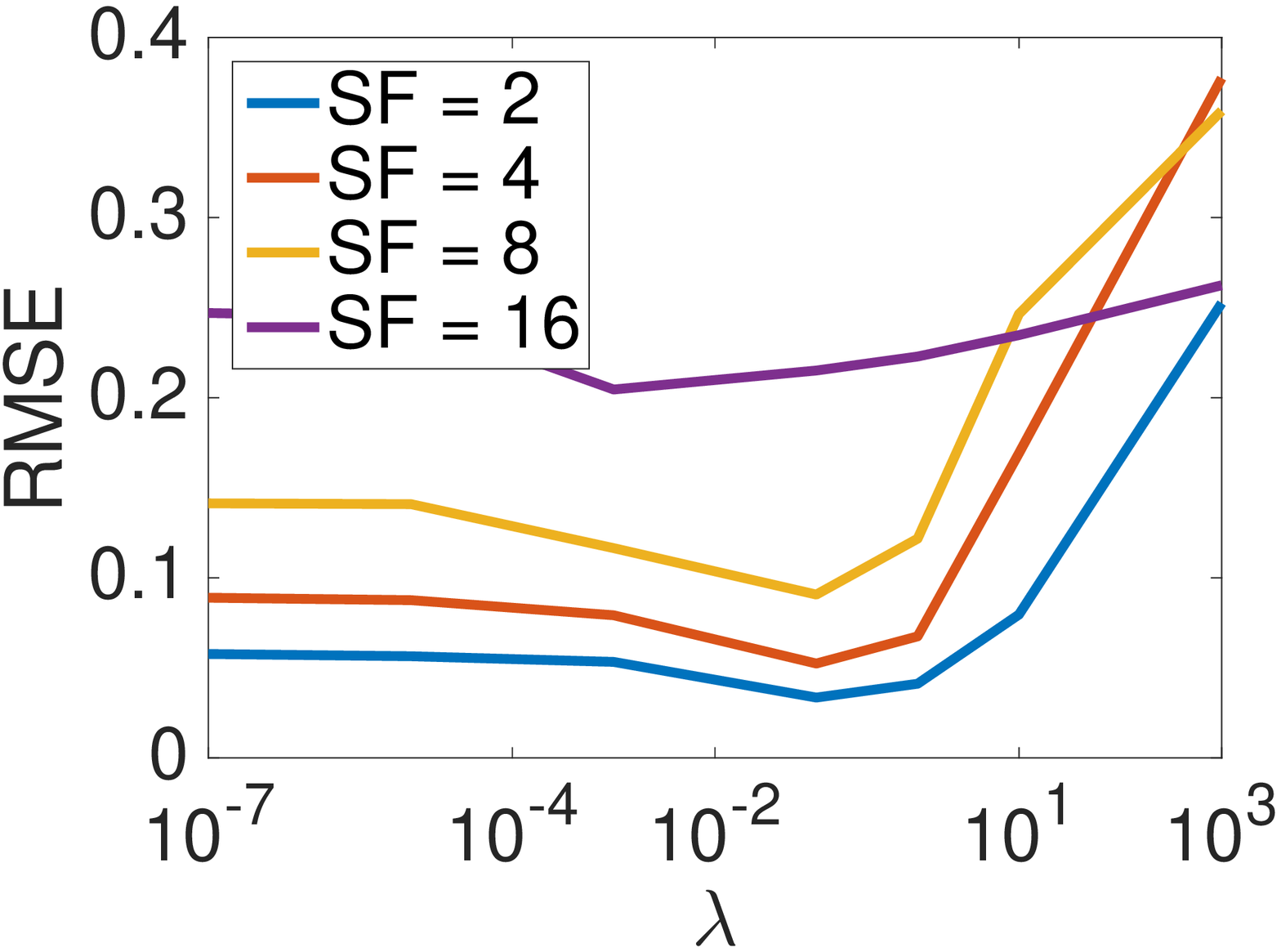} \\ 
   \!\!\!\! {\small (a)} & \!\!\!\!  {\small (b)}             
  \end{tabular}
\caption{(a-b) Impact of the regularization parameter $\lambda$ on accuracy. The interval $\lambda \in [10^{-2},10^2]$ constitutes an appropriate choice which both avoids the ambiguities of uncalibrated photometric stereo and the ill-posedness of super-resolution.  }
\label{fig:4}
\end{figure}

\paragraph{Parameter Tuning --} The only parameter in Model~\eqref{eq:12} is $\lambda$, which controls the respective influence of the super-resolution and photometric terms.  As expected, $\lambda \to 0$ yields a loss of fine-scale details (high mean angular error on normals due to the absence of photometric stereo-based estimation)  while $\lambda \to \infty$ leads to a low-frequency bias (high root mean square error on depth due to the generalized bas-relief ambiguity). Figure~\ref{fig:4} shows that the range $\lambda \in [10^{-2},10^2]$ provides satisfactory results. If not stated otherwise, the value $\lambda = 0.1$ is thus used in all our synthetic experiments.

\begin{figure}[!ht]
  \begin{tabular}{@{\hspace{-0.05em}}cc}
   \!\!\!\! \includegraphics[width = 0.49\linewidth, trim = 0em 0em 2.5em 1em, clip]{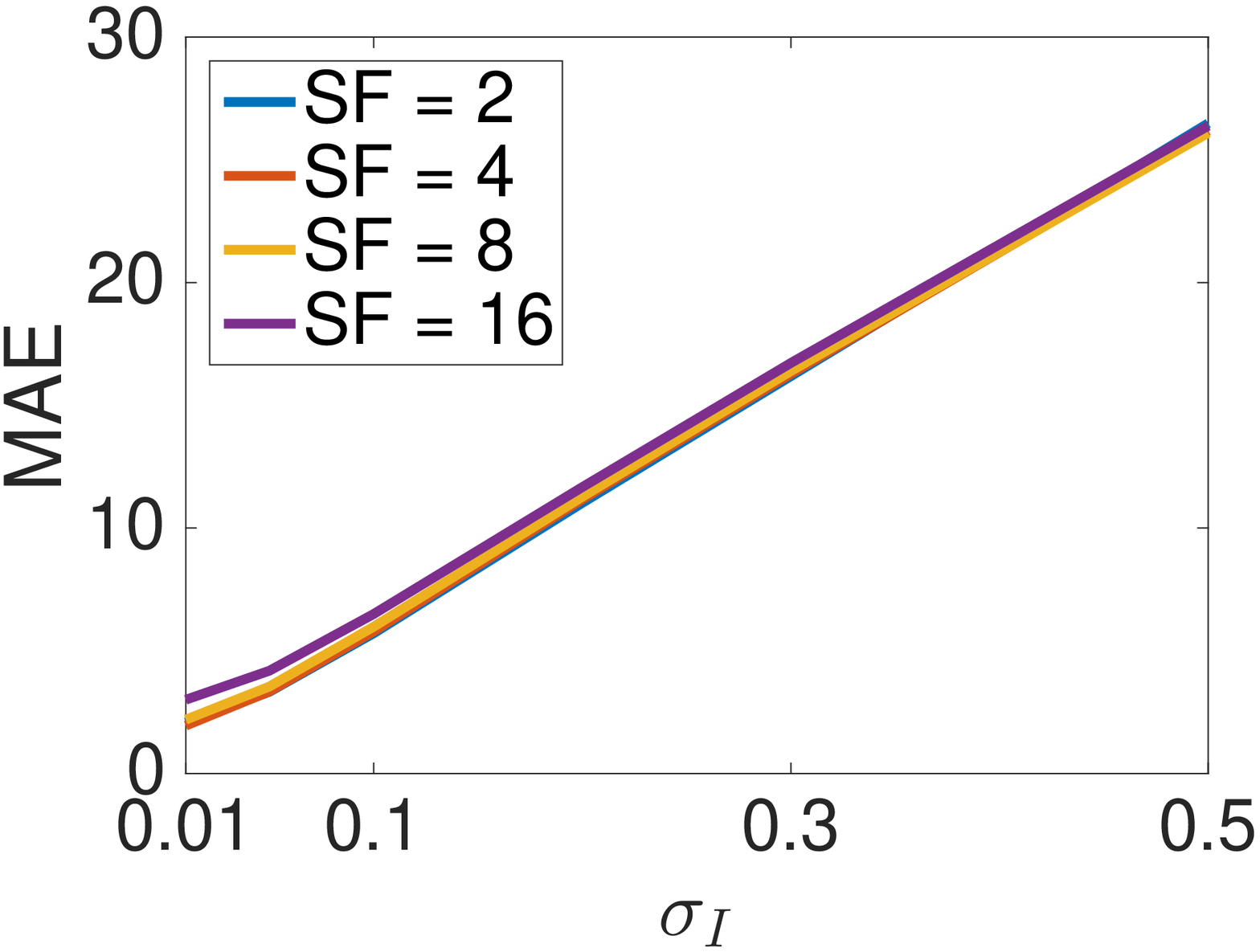} & \!\!\!\! 
    \includegraphics[width = 0.49\linewidth, trim = 0em 0em 2.5em 1em, clip]{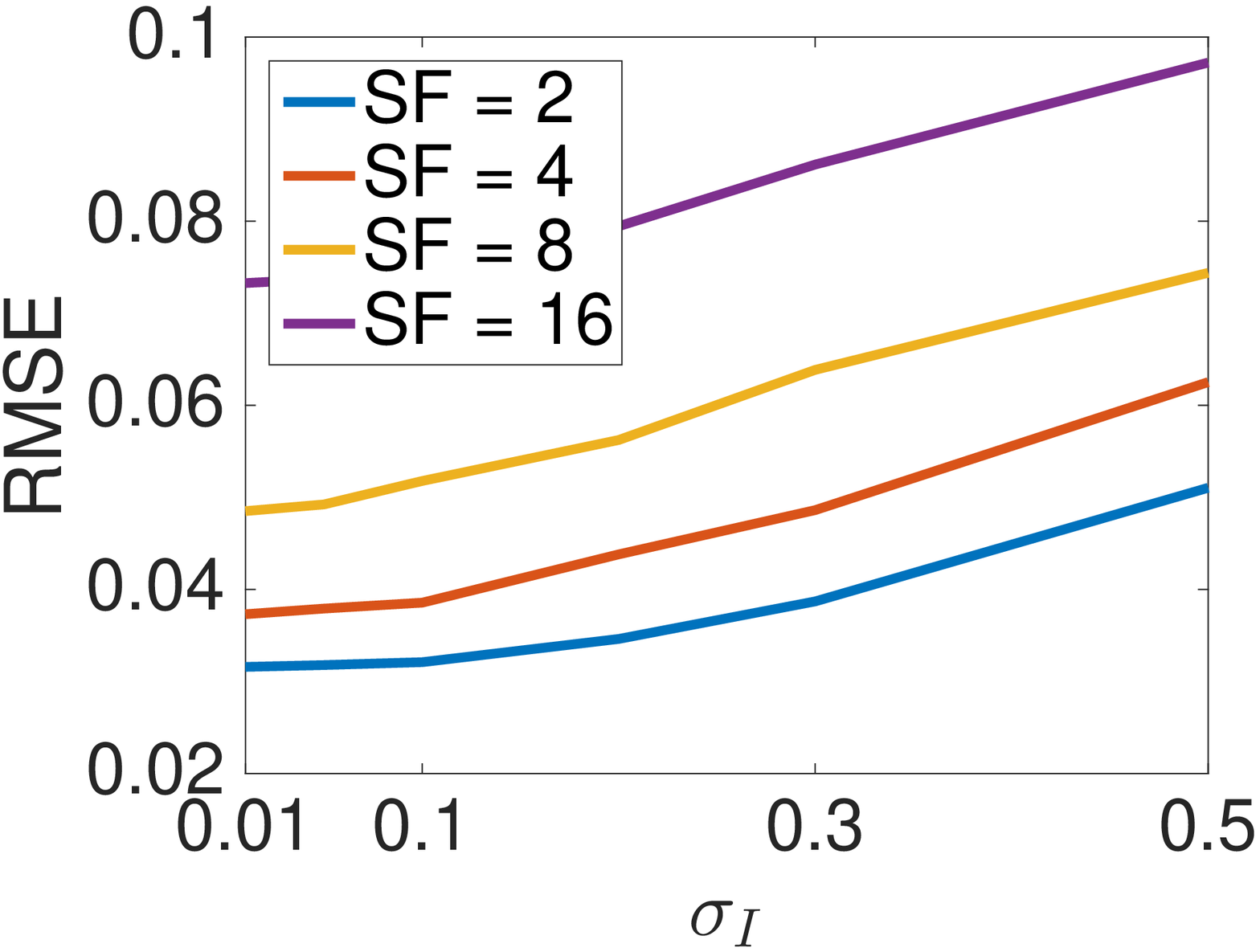} \\ 
   \!\!\!\! {\small (a)} & \!\!\!\!  {\small (b)}  \\
   \!\!\!\! \includegraphics[width = 0.49\linewidth, trim = 0em 0em 2.5em 1em, clip]{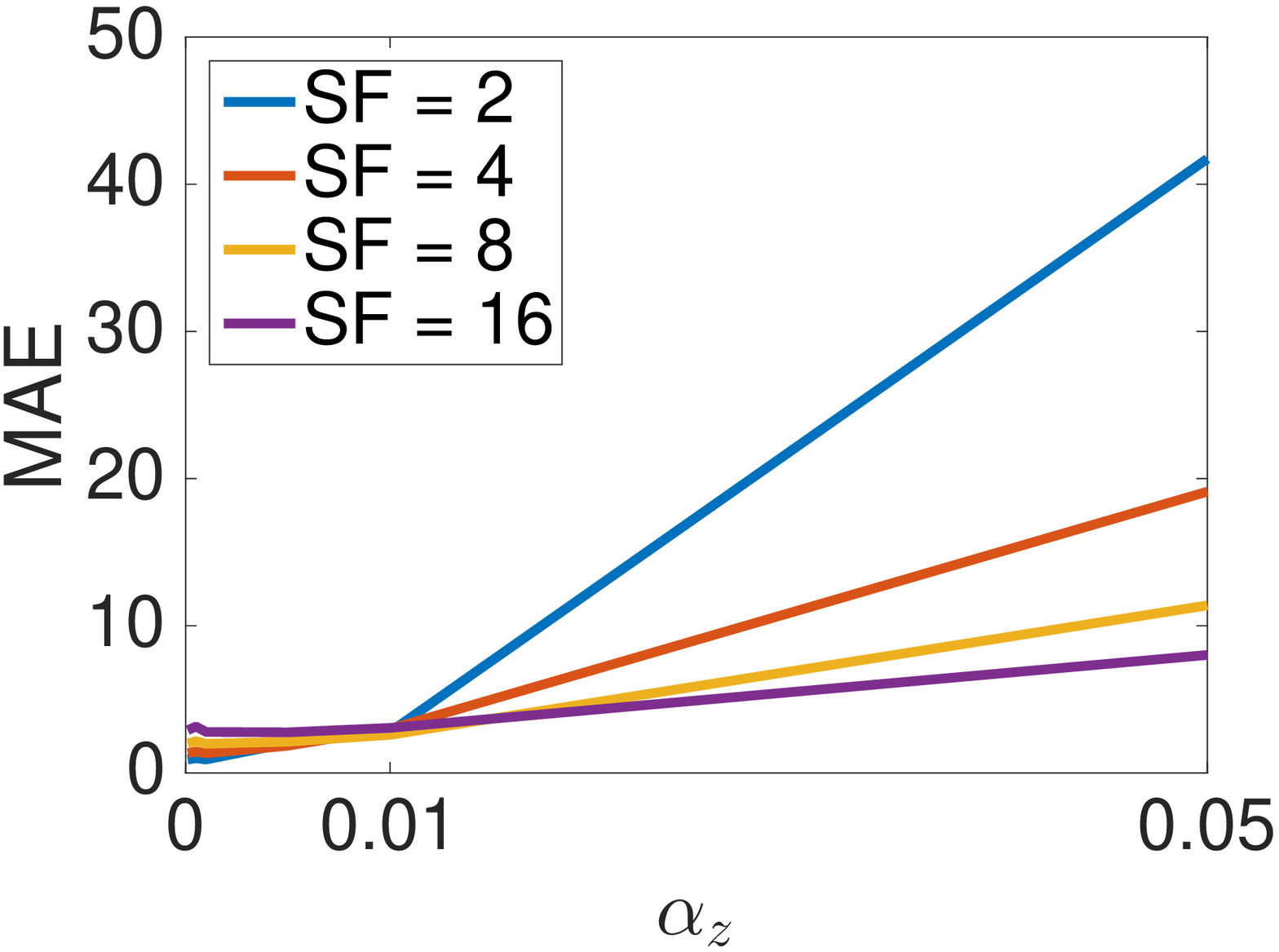} & \!\!\!\! 
    \includegraphics[width = 0.49\linewidth, trim = 0em 0em 2.5em 1em, clip]{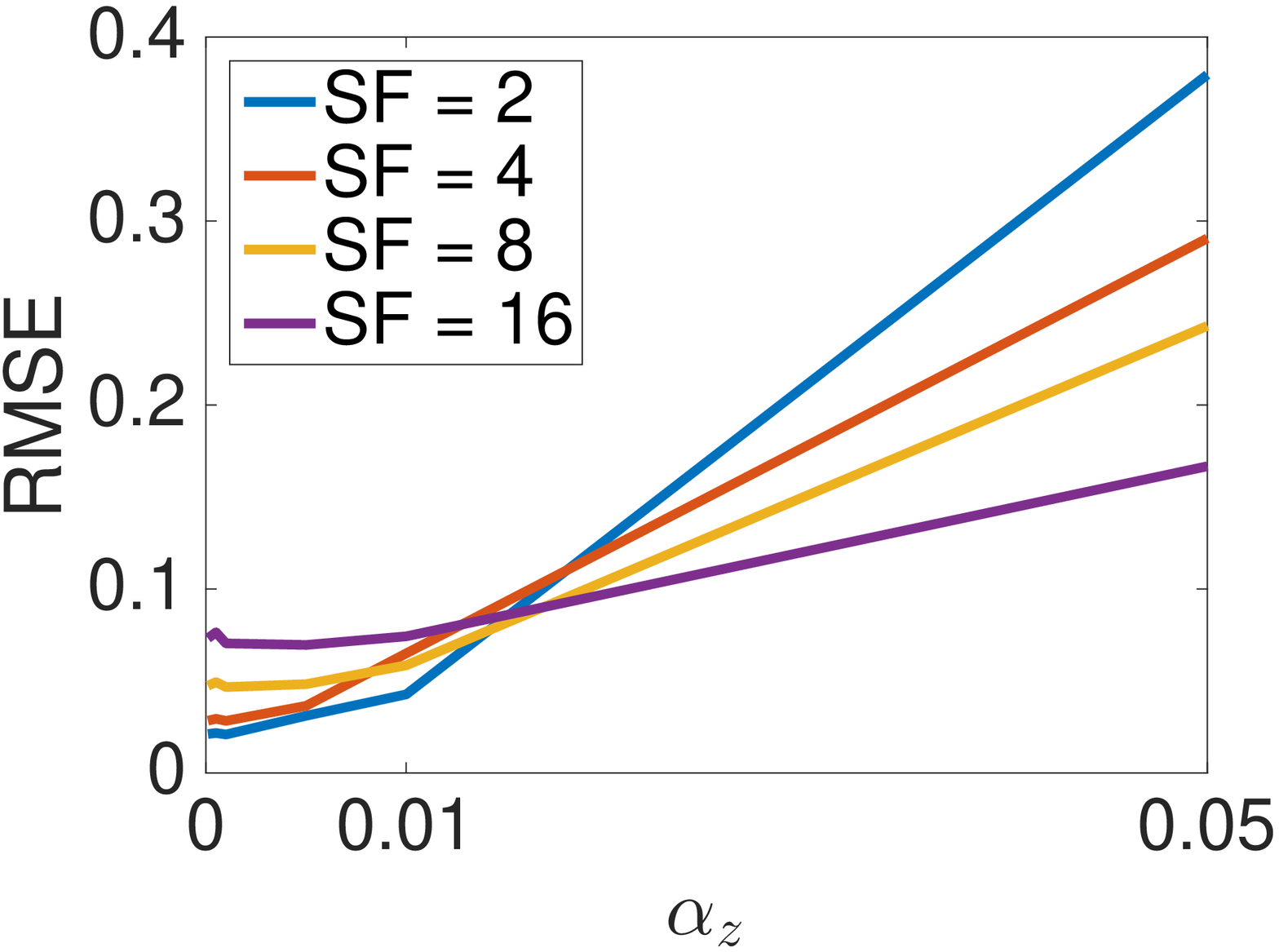} \\ 
   \!\!\!\! {\small (c)} & \!\!\!\!  {\small (d)}  \\            
  \end{tabular}
\caption{(a-b) Impact of the amount of zero-mean, Gaussian noise added to the RGB images on accuracy. (c-d) Same, with increasing amount of noise on the LR depth maps. Accuracy deteriorates linearly with respect to both noise levels. However, typical values for $\alpha_z$ are around $10^{-5}$ in real-world scenarios using a Microsoft Kinect v1~\cite{khoshelham2012}, so our method is more robust to this type of noise than required.}
\label{fig:5}
\end{figure}

\paragraph{Robustness to Noise --} Figure~\ref{fig:5} evaluates robustness to noise in both the input LR depth maps and the HR RGB images. Unsurprisingly, accuracy severely deteriorates if noise is tremendous in both depth and RGB images. However, our approach is robust to a realistic amount of noise. 

\begin{figure}[!ht]
  {\renewcommand{\arraystretch}{1.1}
  \begin{tabular}{@{\hspace{0.0em}}c@{\hspace{0.3em}}c@{\hspace{0.3em}}c@{\hspace{0.3em}}c@{\hspace{0.0em}}}
    \includegraphics[width = 0.49\linewidth, trim = 0em 2em 0em 2em, clip, frame]{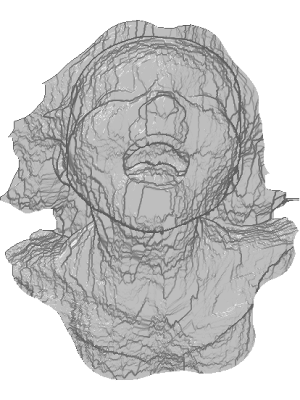}  &  
    \includegraphics[width = 0.49\linewidth, trim = 0em 2em 0em 2em, clip, frame]{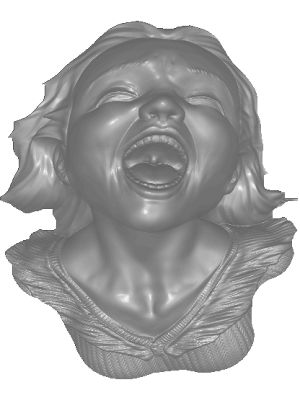}  \\
    {\small RMSE $ = 0.0728$} &  {\small RMSE $ = 0.9199$} \\
    {\small MAE $= 34.4129$} &  {\small MAE $= 41.8041$} \\
    {\small (a)} &  {\small (b)} \\
    &\\
    \includegraphics[width = 0.26\linewidth, trim = 0em 2em 0em 2em, clip, frame]{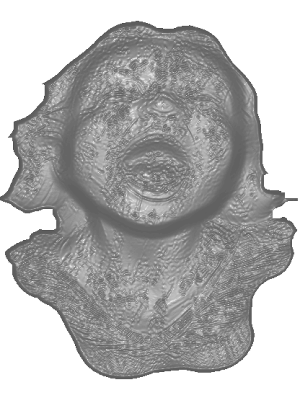}  &  
    \includegraphics[width = 0.49\linewidth, trim = 0em 2em 0em 2em, clip, frame]{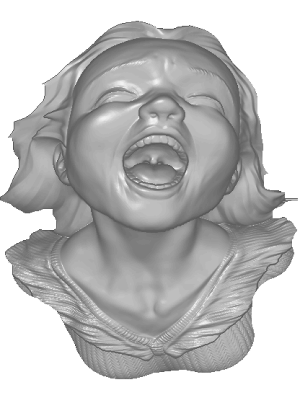}  \\
    {\small RMSE $ = 0.1655$} &  {\small RMSE $ = 0.03139 $} \\
    {\small MAE $= 38.9316$} &  {\small MAE $=1.4528$} \\
     {\small (c)} &  {\small (d)}\\ 
  \end{tabular}}
\caption{Comparison between (a) image-driven depth super-resolution using \eqref{eq:2} with \eqref{eq:3}, (b) uncalibrated photometric stereo~\cite{Papadhimitri2014b}, (c) single-image RGB-D fusion~\cite{Or-el2015} and (d) the proposed photometric stereo-aware super-resolution method. Image-based depth super-resolution and RGB-D fusion are unable to appropriately handle spatially-varying reflectance. Uncalibrated photometric appropriately estimates reflectance and thin structures, but it is prone to a high low-frequency bias due to the generalized bas-relief ambiguity. The proposed photometric stereo-aware super-resolution circumvents both these issues. For the input depth map the RMSE is $0.0579$ and MAE is $65.7150$. }
\label{fig:6}
\end{figure}

\paragraph{Comparison with Other Methods --} Eventually, Figure~\ref{fig:6} shows the advantage of our approach over standard image-driven depth super-resolution, pure uncalibrated photometric stereo, and shading-based refinement using a single LR RGB image. In this experiment, the depth super-resolution approach was implemented using \eqref{eq:2} with \eqref{eq:3} being the corresponding regularization term. Image-driven super-resolution interprets sharp image discontinuities as sharp depth features, because it is not able to estimate reflectance. For uncalibrated photometric stereo, we employed code from the authors of~\cite{Papadhimitri2014b}. This method is able to estimate the albedo, but it still requires a prior in order to solve the generalized bas-relief ambiguity. It is thus not purely data-driven and subject to bias.
As for shading-based refinement, the RGB-D fusion code provided in~\cite{Or-el2015} was used. It assumes that both the RGB image and the depth map have the same resolution, thus it does not achieve super-resolution. Still, this experiment highlights the advantage of using a multiple-light setup: shape-from-shading methods have to introduce a smoothness prior on the reflectance, which is often non-realistic and induces artefacts on the depth around reflectance discontinuities. Only by controlling the lighting this prior can be avoided, and overall the proposed method, which both estimates reflectance and solves the photometric ambiguities, provides the best results.

\begin{figure*}[!ht]
  \begin{tabular}{cccccc}
    \includegraphics[width = 0.14\linewidth]{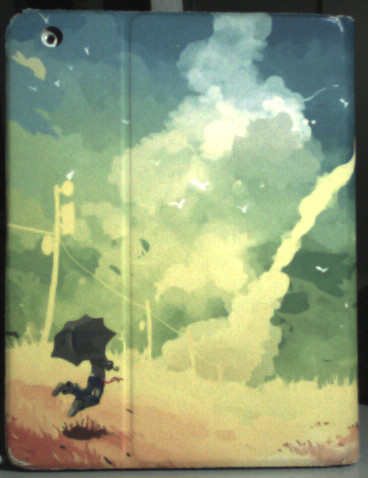}  &
    \includegraphics[width = 0.14\linewidth]{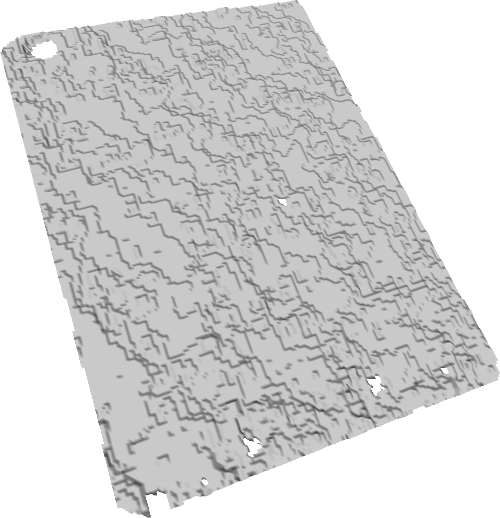} & 
    \!\!\!\!\!\!\!\!\!\!\!\!\! \includegraphics[width = 0.19\linewidth]{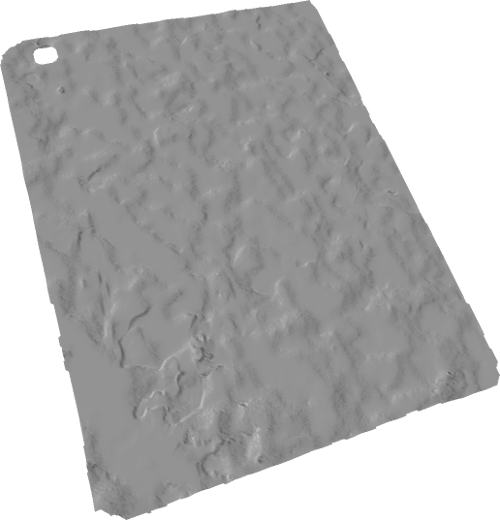}  &  
    \!\!\!\!\!\!\! \includegraphics[width = 0.19\linewidth]{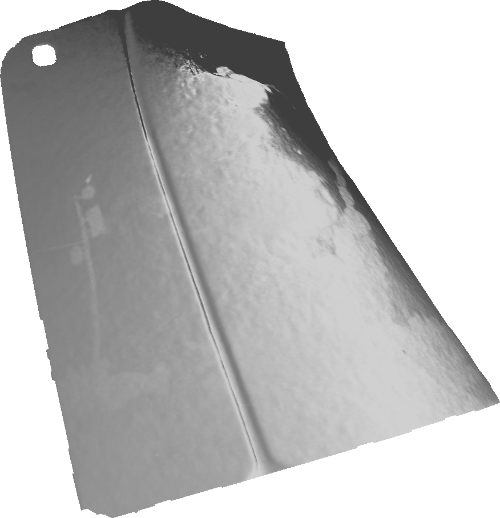}  &  
    \!\!\!\!\!\!\!\!\! \includegraphics[width = 0.14\linewidth]{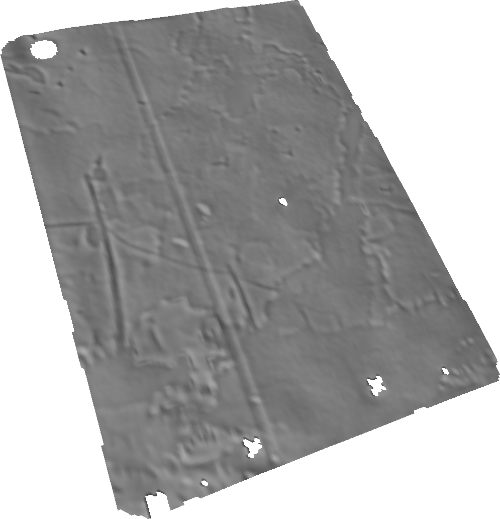} & 
    \!\!\!\!\!\!\!\!\!\!\!\!\! \includegraphics[width = 0.19\linewidth]{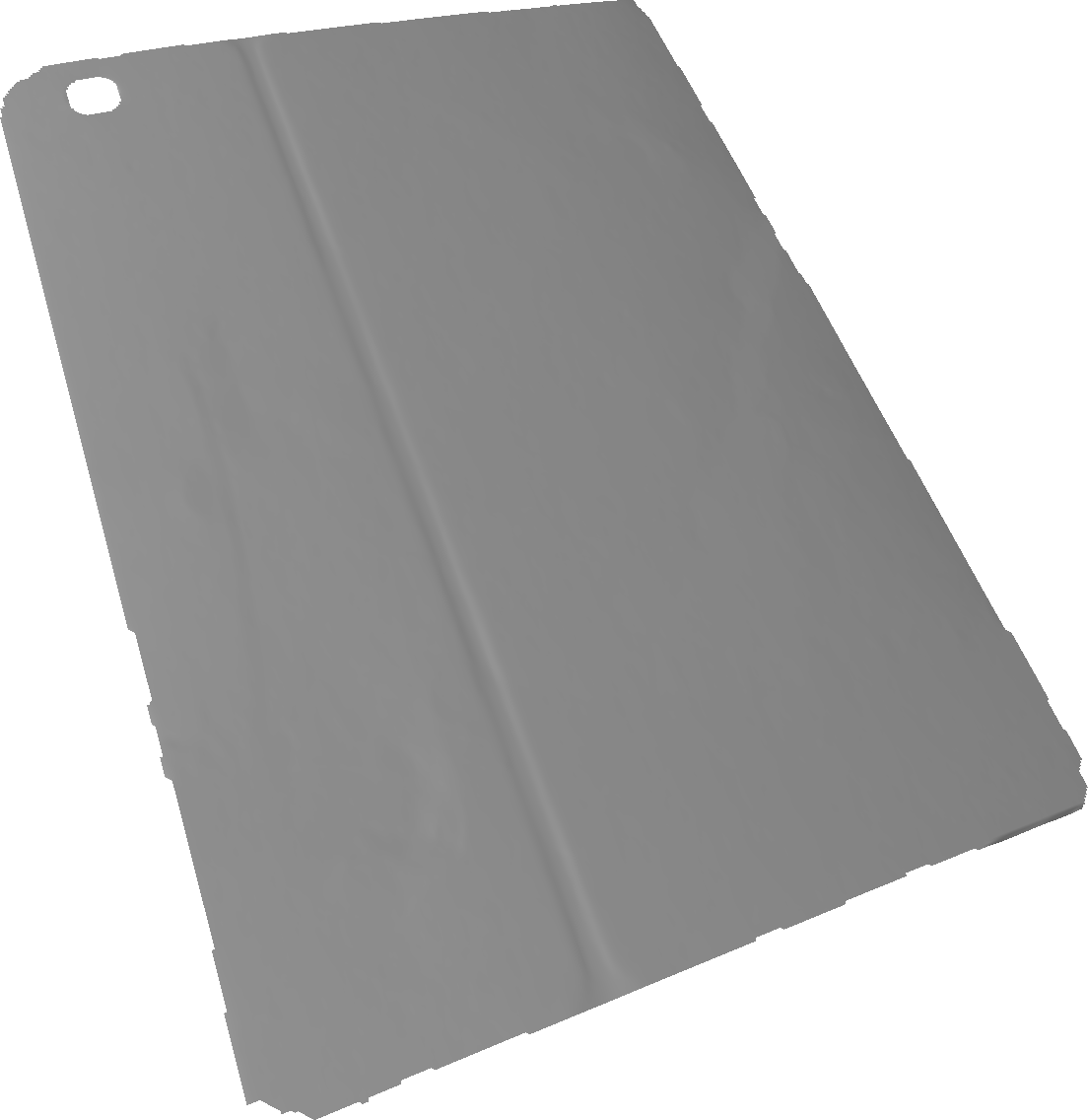}  \\
    
    \includegraphics[width = 0.12\linewidth]{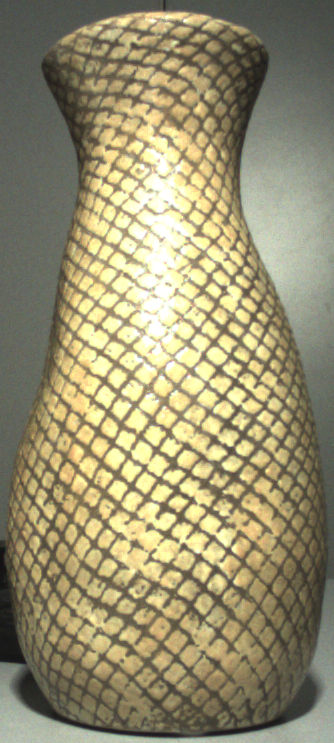}  &
    \includegraphics[width = 0.065\linewidth]{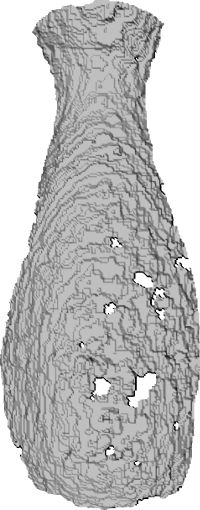} & 
    \includegraphics[width = 0.105\linewidth]{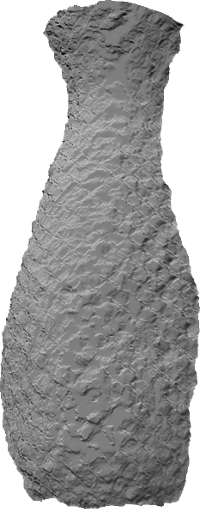} & 
    \includegraphics[width = 0.1\linewidth]{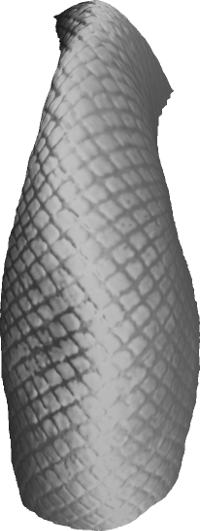} & 
    \includegraphics[width = 0.065\linewidth]{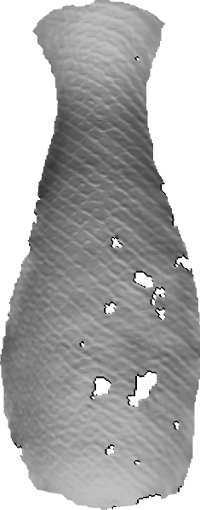} & 
    \includegraphics[width = 0.1\linewidth]{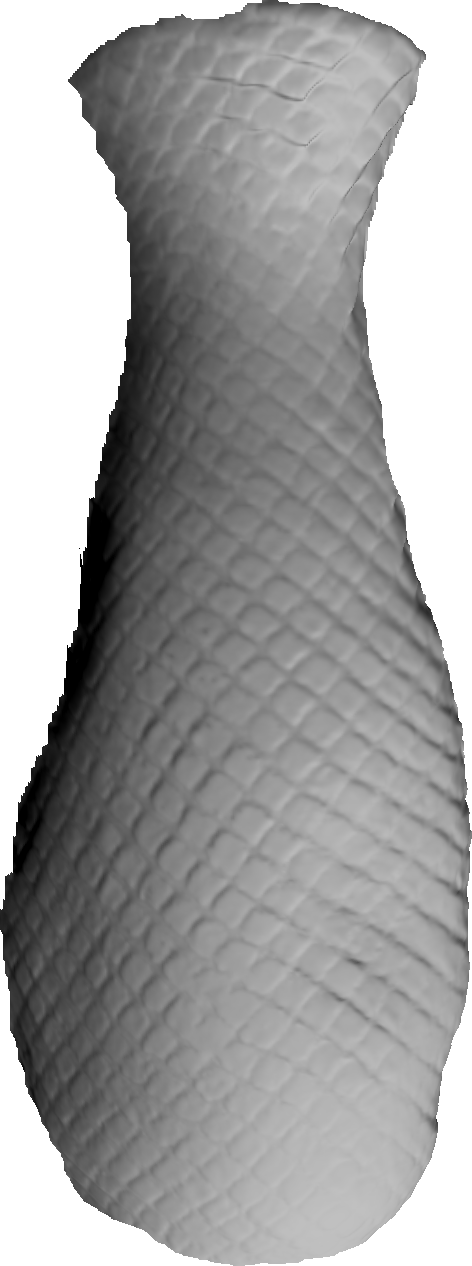} \\            
    
    {\small (a)} &  {\small (b)} & {\small (c)} &  {\small (d)} &  {\small (e)} &  {\small (f)} 
  \end{tabular}
\caption{Comparison between the proposed method and others, on real-world datasets. (a) One (out of $n=20$) HR RGB image. (b) One of the LR depth maps (SF $= 2$). (c) Image-based depth super-resolution (Equations~\eqref{eq:2} and~\eqref{eq:3}). (d) Uncalibrated photometric stereo~\cite{Papadhimitri2014b}. (e) RGB-D fusion~\cite{Or-el2015}. (f) Proposed method ($\lambda = 1$). These results confirm the conclusion of the synthetic experiments in Figure~\ref{fig:6}.  }
\label{fig:7}
\end{figure*}

\begin{figure*}[!ht]
\centering
  \begin{tabular}{ccccc}
    \includegraphics[width = 0.2\linewidth]{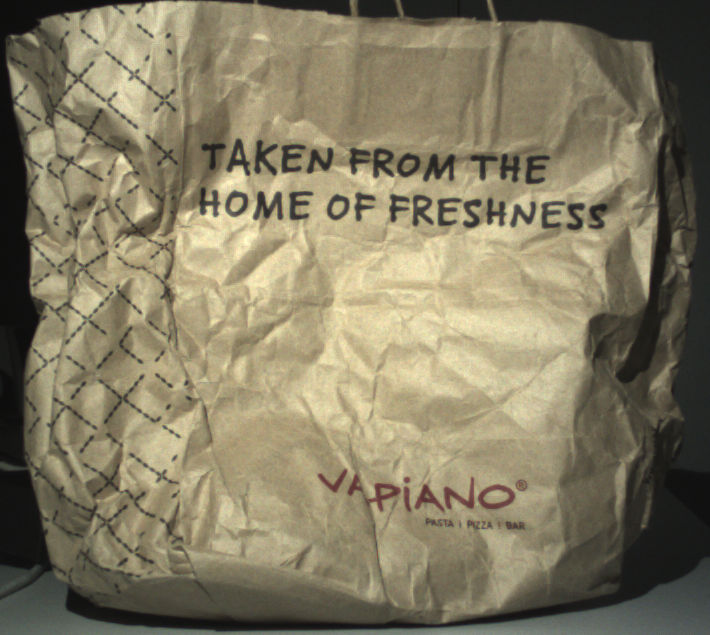}  &
    \includegraphics[width = 0.10\linewidth]{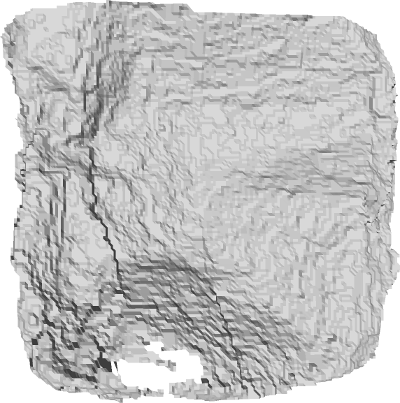} & 
    \includegraphics[width = 0.165\linewidth]{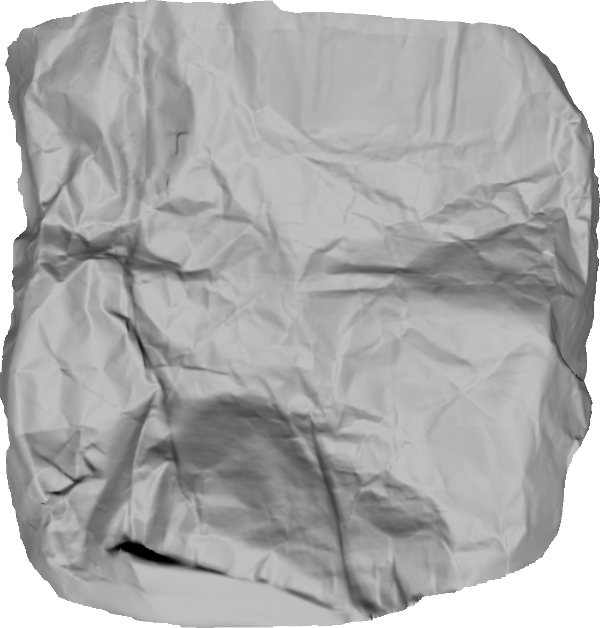}  &  
    \includegraphics[width = 0.2\linewidth]{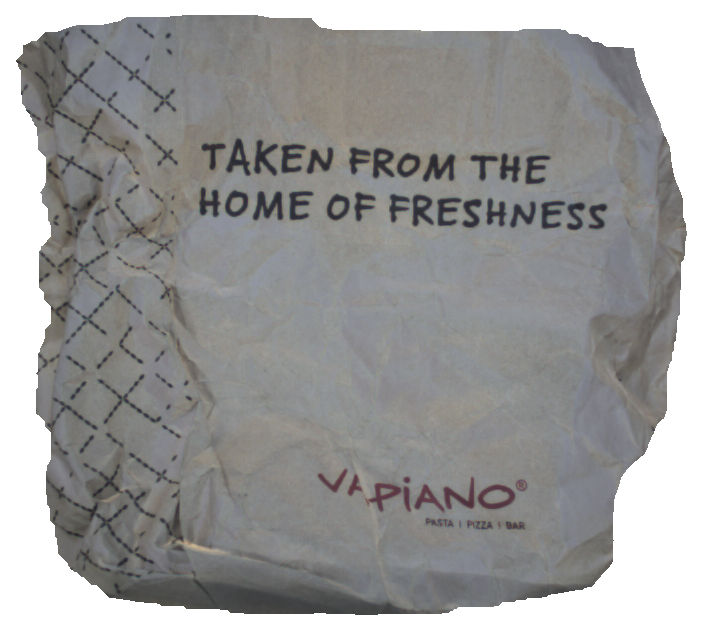} &
    \includegraphics[width = 0.22\linewidth]{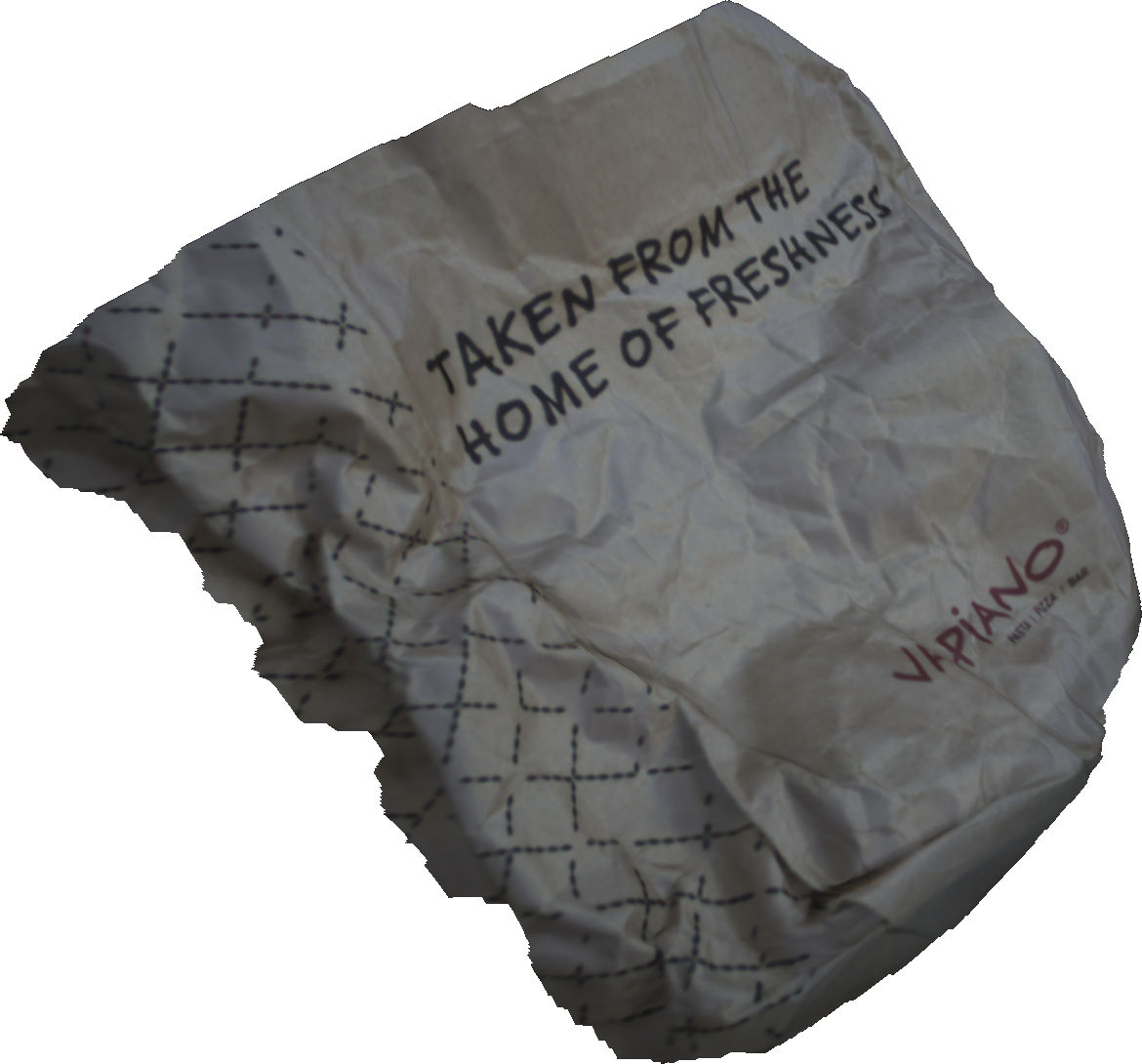} \\
    \includegraphics[width = 0.175\linewidth]{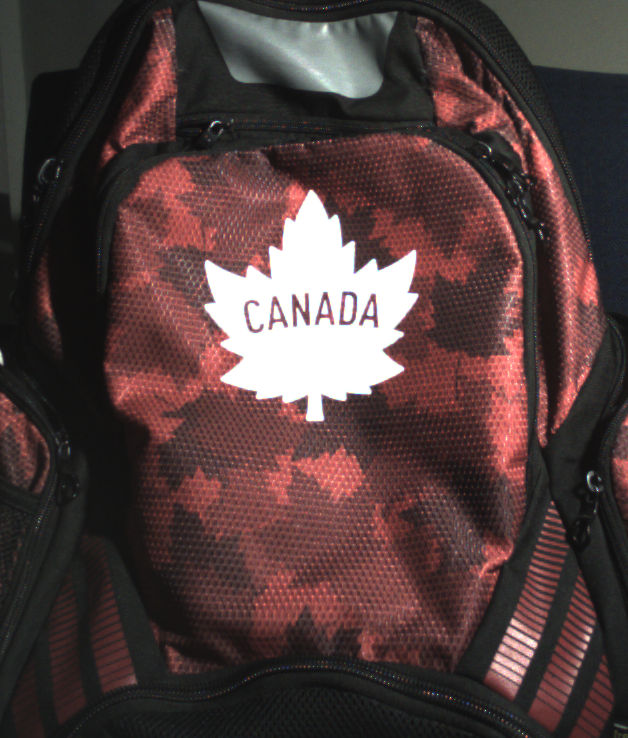}  &
    \includegraphics[width = 0.1\linewidth]{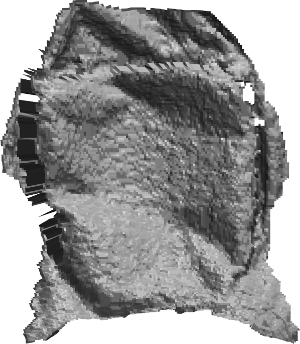} & 
    \includegraphics[width = 0.182\linewidth]{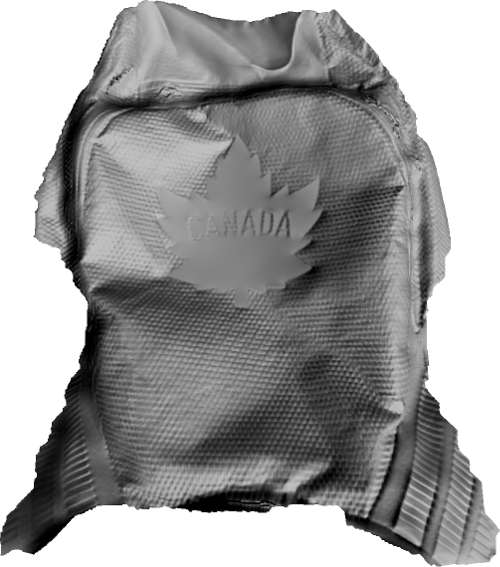}  &  
    \includegraphics[width = 0.175\linewidth]{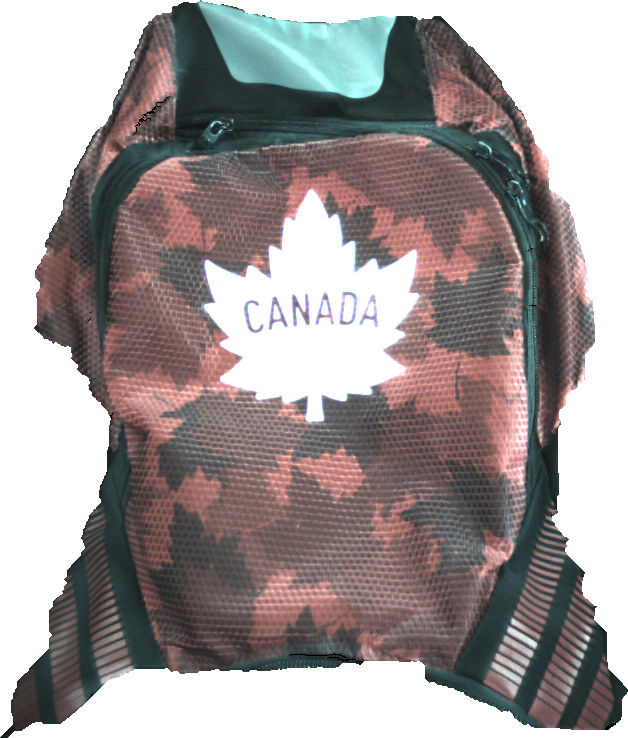} &
    \includegraphics[width = 0.21\linewidth]{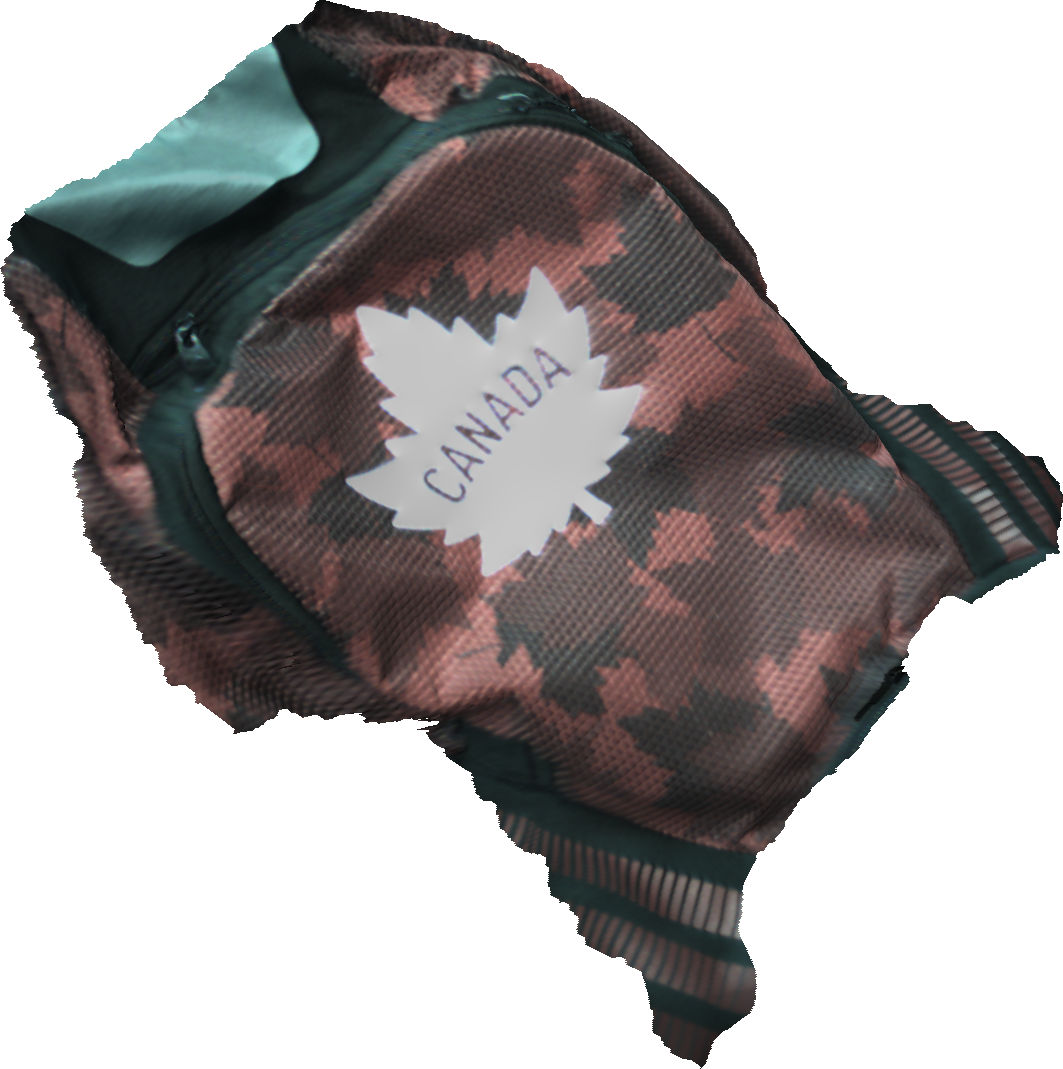}  \\

    {\small (a)} &  {\small (b)} & {\small (c)} &  {\small (d)} &  {\small (e)}
  \end{tabular}
\caption{Qualitative results on real-world datasets. (a) One of the HR RGB images. (b) One of the LR depth maps (SF $= 4$, but the LR depth maps are enlarged for the sake of illustration). (c) Estimated HR depth map (paper bag $\lambda=40$, backpack $\lambda=10$). (d) Estimated HR reflectance map. (e) Relighting of the HR 3D-model from new viewing and lighting angles.}
\label{fig:8}
\end{figure*}

\subsection{Qualitative Evaluation on Real-world Datasets}

For real-world experiments, we use the Asus Xtion Pro Live, which has the same depth sensor as Microsoft's Kinect v1. The sensors provide a maximum RGB resolution of $1280 \times 1024$ $px$ and QVGA ($320 \times 240$ $px$) or VGA ($640 \times 480$ $px$) depth resolution. Data is acquired in video mode, while moving a single white Luxeon Rebel LED in front of the object. Experiments are run in an office room with ambient lighting. From each sequence, we extract a series of $n=20$ LR depth maps and HR RGB images. From the user perspective, acquisition of data is thus extremely simple,  since no calibration is required.

We consider in Figures~\ref{fig:teaser},~\ref{fig:7} and~\ref{fig:8} five different objects: a shirt with piecewise-constant reflectance, a tablet cover with piecewise-smooth reflectance and a fine wrinkle, a partly specular vase, a creased bag with some text painting, and a backpack with very thin geometric structures and piecewise-constant reflectance. 

In all these cases, our method is able to successfully upsample the depth maps, while also recovering the fine geometric structures and separate reflectance from shading. Interestingly, robustness to specularities is enforced, although we only model Lambertian reflectance. This is probably due to having a rough prior on shape.

\newpage
~\\
\newpage

\section{Conclusion}
\label{sec:5} 

We have presented a novel variational framework for depth super-resolution in RGB-D sensing, by resorting to the photometric stereo technique. For this task, it is enough to capture a sequence of low-resolution depth maps and high-resolution RGB images under uncalibrated, varying illumination. Then, the proposed variational framework is able to carry out unambiguous shape, reflectance and lighting estimation. The low-resolution depth measurements essentially disambiguate uncalibrated photometric stereo and, symmetrically, the photometric stereo-based regularization term disambiguates super-resolution. The proposed method can be used out-of-the-box using common devices, without any need for calibration. This is made possible by the tailored photometric stereo regularizer which implicitly ensures regularity of the super-resolved depth map. 

For the future work, we will explore with more care the theoretical foundations of the proposed variational framework, and prove uniqueness of the solution by resorting to a continuous analysis of the problem. 

{\small
\bibliographystyle{ieee}
\bibliography{biblio_Yvain.bib,biblio_Songyou.bib,biblio_Bjorn.bib}

\begin{thebibliography}{10}\itemsep=-1pt

\bibitem{Alldrin2007}
N.~G. Alldrin, S.~P. Mallick, and D.~J. Kriegman.
\newblock {Resolving the generalized bas-relief ambiguity by entropy
  minimization}.
\newblock In {\em CVPR}, 2007.

\bibitem{Anderson2011}
R.~Anderson, B.~Stenger, and R.~Cipolla.
\newblock Augmenting depth camera output using photometric stereo.
\newblock In {\em MVA}, 2011.

\bibitem{Basri2007}
R.~Basri, D.~W. Jacobs, and I.~Kemelmacher.
\newblock {Photometric stereo with general, unknown lighting}.
\newblock {\em IJCV}, 72(3):239--257, 2007.

\bibitem{Belhumeur1999}
P.~N. Belhumeur, D.~J. Kriegman, and A.~L. Yuille.
\newblock {The bas-relief ambiguity}.
\newblock {\em IJCV}, 35(1):33--44, 1999.

\bibitem{Chatterjee2015}
A.~Chatterjee and V.~Madhav~Govindu.
\newblock Photometric refinement of depth maps for multi-albedo objects.
\newblock In {\em CVPR}, 2015.

\bibitem{Chaudhuri2005}
S.~Chaudhuri and M.~V. Joshi.
\newblock {\em Motion-free super-resolution}.
\newblock Springer Verlag, 2005.

\bibitem{Cristani2004}
M.~Cristani, D.~S. Cheng, V.~Murino, and D.~Pannullo.
\newblock Distilling information with super-resolution for video surveillance.
\newblock In {\em VSSN}, 2004.

\bibitem{Fablet2015}
R.~Fablet and F.~Rousseau.
\newblock Missing data super-resolution using non-local and statistical priors.
\newblock In {\em ICIP}, 2015.

\bibitem{Goldlucke2014}
B.~Goldl{\"u}cke, M.~Aubry, K.~Kolev, and D.~Cremers.
\newblock A super-resolution framework for high-accuracy multiview
  reconstruction.
\newblock {\em IJCV}, 106(2):172--191, 2014.

\bibitem{Greenspan2008}
H.~Greenspan.
\newblock Super-resolution in medical imaging.
\newblock {\em The Computer Journal}, 52(1):43--63, 2008.

\bibitem{Han2013}
Y.~Han, J.-Y. Lee, and I.~So~Kweon.
\newblock High quality shape from a single rgb-d image under uncalibrated
  natural illumination.
\newblock In {\em ICCV}, 2013.

\bibitem{Hayakawa1994}
H.~Hayakawa.
\newblock {Photometric stereo under a light source with arbitrary motion}.
\newblock {\em JOSA A}, 11(11):3079--3089, 1994.

\bibitem{Horn1989}
B.~K.~P. Horn and M.~J. Brooks, editors.
\newblock {\em {Shape from shading}}.
\newblock MIT Press, 1989.

\bibitem{khoshelham2012}
K.~Khoshelham and S.~O. Elberink.
\newblock Accuracy and resolution of kinect depth data for indoor mapping
  applications.
\newblock {\em Sensors}, 12(2):1437--1454, 2012.

\bibitem{Kim2015}
K.~Kim, A.~Torii, and M.~Okutomi.
\newblock Joint estimation of depth, reflectance and illumination for depth
  refinement.
\newblock In {\em ICCVW}, 2015.

\bibitem{Lu2017}
F.~Lu, X.~Chen, I.~Sato, and Y.~Sato.
\newblock {SymPS: BRDF symmetry guided photometric stereo for shape and light
  source estimation}.
\newblock {\em PAMI}, (to appear), 2017.

\bibitem{Lu2013}
Z.~Lu, Y.-W. Tai, F.~Deng, M.~Ben-Ezra, and M.~S. Brown.
\newblock {A 3D imaging framework based on high-resolution photometric-stereo
  and low-resolution depth}.
\newblock {\em IJCV}, 102(1-3):18--32, 2013.

\bibitem{Maier2015}
R.~Maier, J.~St{\"u}ckler, and D.~Cremers.
\newblock {Super-resolution keyframe fusion for 3D modeling with high-quality
  textures}.
\newblock In {\em 3DV}, 2015.

\bibitem{Marquina2008}
A.~Marquina and S.~J. Osher.
\newblock {Image super-resolution by TV-regularization and Bregman iteration}.
\newblock {\em J. Sci. Comput.}, 37(3):367--382, 2008.

\bibitem{Or-el2015}
R.~Or-el, G.~Rosman, A.~Wetzler, R.~Kimmel, and A.~M. Bruckstein.
\newblock {RGBD-fusion: real-time high precision depth recovery}.
\newblock In {\em {CVPR}}, 2015.

\bibitem{Papadhimitri2014b}
T.~Papadhimitri and P.~Favaro.
\newblock {A closed-form, consistent and robust solution to uncalibrated
  photometric stereo via local diffuse reflectance maxima}.
\newblock {\em IJCV}, 107(2):139--154, 2014.

\bibitem{Park2014}
J.~Park, H.~Kim, Y.~W. Tai, M.~S. Brown, and I.~S. Kweon.
\newblock High-quality depth map upsampling and completion for rgb-d cameras.
\newblock {\em TIP}, 23(12):5559--5572, 2014.

\bibitem{Queau2017}
Y.~Qu\'eau, T.~Wu, F.~Lauze, J.-D. Durou, and D.~Cremers.
\newblock A non-convex variational approach to photometric stereo under
  inaccurate lighting.
\newblock In {\em CVPR}, 2017.

\bibitem{Tan2008}
P.~Tan, S.~Lin, and L.~Quan.
\newblock Subpixel photometric stereo.
\newblock {\em PAMI}, 30(8):1460--1471, 2008.

\bibitem{Tsai1984}
R.~Y. Tsai and T.~S. Huang.
\newblock Multiframe image restoration and registration.
\newblock {\em Advances in Computer Vision and Image Processing},
  1(2):317--339, 1984.

\bibitem{unger2010}
M.~Unger, T.~Pock, M.~Werlberger, and H.~Bischof.
\newblock A convex approach for variational super-resolution.
\newblock In {\em DAGM}, 2010.

\bibitem{ouwerkerk2006}
J.~D. Van~Ouwerkerk.
\newblock Image super-resolution survey.
\newblock {\em Image and vision Computing}, 24(10):1039--1052, 2006.

\bibitem{werlberger2009}
M.~Werlberger, W.~Trobin, T.~Pock, A.~Wedel, D.~Cremers, and H.~Bischof.
\newblock Anisotropic huber-l1 optical flow.
\newblock In {\em BMVC}, 2009.

\bibitem{Woodham1980}
R.~J. Woodham.
\newblock {Photometric method for determining surface orientation from multiple
  images}.
\newblock {\em Opt. Eng.}, 19(1):139--144, 1980.

\bibitem{Wu2014}
C.~Wu, M.~Zollh{\"o}fer, M.~Nie{\ss}ner, M.~Stamminger, S.~Izadi, and
  C.~Theobalt.
\newblock Real-time shading-based refinement for consumer depth cameras.
\newblock {\em TOG}, 33(6), 2014.

\bibitem{Yu2013}
L.-F. Yu, S.-K. Yeung, Y.-W. Tai, and S.~Lin.
\newblock Shading-based shape refinement of rgb-d images.
\newblock In {\em CVPR}, 2013.

\bibitem{Yuille1997}
A.~L. Yuille and D.~Snow.
\newblock {Shape and albedo from multiple images using integrability}.
\newblock In {\em CVPR}, 1997.

\end{thebibliography}
}

\end{document}